\documentclass[sigplan,review=false]{acmart}
\renewcommand\footnotetextcopyrightpermission[1]{}
\AtBeginDocument{%
  }

\usepackage{graphicx}
\usepackage{float}
\usepackage{subfig}
\usepackage{booktabs}
\usepackage{multirow}
\usepackage[ruled,linesnumbered]{algorithm2e}
\newcommand{\myparagraph}[1]{\textbf{#1.}}
\usepackage{enumitem}
\setlist[itemize]{topsep=0pt, partopsep=0pt, parsep=0pt}
\usepackage{titlesec}
\titlespacing*{\section}{0pt}{1pt}{1pt}
\titlespacing*{\subsection}{0pt}{1pt}{1pt}
\titlespacing*{\subsubsection}{0pt}{1pt}{1pt}
\newcommand{\xxx}{Hybrid-Parallel}
\newcommand{\xxxparallel}{LOP}
\newcommand{\xxxschedule}{LOSS}
\newcommand{\github}{\url{https://github.com/SunZekai-CN/HybridParallel}}
\begin{document}

\title{\xxx: Achieving High Performance and Energy Efficient Distributed Inference on Robots}

\author{Zekai Sun}
\orcid{0000-0003-0269-7940}
\affiliation{%
  \institution{The University of Hong Kong}
  \city{Hong Kong}
  \country{China}
}
\email{zksun@cs.hku.hk}

\author{Xiuxian Guan}
\orcid{0000-0001-6133-8388}
\affiliation{%
  \institution{The University of Hong Kong}
  \city{Hong Kong}
  \country{China}
}
\email{xxguan@cs.hku.hk}

\author{Junming Wang}
\orcid{0000-0002-2271-8270}
\affiliation{%
  \institution{The University of Hong Kong}
  \city{Hong Kong}
  \country{China}
}
\email{jmwang@cs.hku.hk}

\author{Haoze Song}
\affiliation{%
  \institution{The University of Hong Kong}
  \city{Hong Kong}
  \country{China}
}
\email{hzsong@cs.hku.hk}

\author{Yuhao Qing}
\affiliation{%
  \institution{The University of Hong Kong}
  \city{Hong Kong}
  \country{China}
}
\email{yhqing@cs.hku.hk}

\author{Tianxiang Shen}
\orcid{0000-0002-6116-5488}
\affiliation{%
  \institution{The University of Hong Kong}
  \city{Hong Kong}
  \country{China}
}
\email{txshen@cs.hku.hk}

\author{Dong Huang}
\affiliation{%
  \institution{The University of Hong Kong}
  \city{Hong Kong}
  \country{China}
}
\email{jdhuang@cs.hku.hk}

\author{Fangming Liu}
\orcid{0000-0002-8570-1345}
\affiliation{%
  \institution{Peng Cheng Laboratory, and Huazhong University of Science and Technology}
  \city{Wuhan}
  \country{China}
}
\email{fmliu@hust.edu.cn}

\author{Heming Cui}
\authornote{Heming Cui is the corresponding author.}
\orcid{0000-0001-7746-440X}
\affiliation{%
  \institution{The University of Hong Kong}
  \city{Hong Kong}
  \country{China}
}
\email{heming@cs.hku.hk}


\begin{abstract}
The rapid advancements in machine learning techniques have led to significant achievements in various real-world robotic tasks. 
These tasks heavily rely on fast and energy-efficient inference of deep neural network (DNN) models when deployed on robots.
To enhance inference performance, distributed inference has emerged as a promising approach, parallelizing inference across multiple powerful GPU devices in modern data centers using techniques such as data parallelism, tensor parallelism, and pipeline parallelism. 
However, when deployed on real-world robots, existing parallel methods fail to provide low inference latency and meet the energy requirements due to the limited bandwidth of robotic IoT.

We present \xxx, a high-performance distributed inference system optimized for robotic IoT. 
\xxx{} employs a fine-grained approach to parallelize inference at the granularity of local operators within DNN layers (i.e., operators that can be computed independently with the partial input, such as the convolution kernel in the convolution layer). 
By doing so, \xxx{} enables different operators of different layers to be computed and transmitted concurrently, and overlap the computation and transmission phases within the same inference task. 
The evaluation demonstrate that \xxx{} reduces inference time by 14.9\% \textasciitilde 41.1\% and energy consumption per inference by up to 35.3\% compared to the state-of-the-art baselines.
\end{abstract}

\maketitle
\section{Introduction}
\label{sec:intro}
The rapid progress in machine learning (ML) techniques has led to remarkable achievements in various fundamental robotic tasks, such as object detection~\cite{Joseph_2021_CVPR,Liu_2022_CVPR,kapao}, robotic control~\cite{li2020graph,agrnav,yang2020multi}, and environmental perception~\cite{cao2022monoscene,li2023voxformer,xia2023scpnet}. 
However, deploying these ML applications on real-world robots requires fast and energy-efficient inference of their deep neural network (DNN) models, given the need for swift environmental responses and the limited battery capacity of robots. 
Placing the entire model on robots not only requires additional computing accelerators on robots (e.g., GPU~\cite{jetsonnx}, FPGA~\cite{ohkawa2018fpga}, SoC~\cite{honkote20192}), but also introduce additional energy consumption (e.g., 162\% more for ~\cite{kapao} in our experiments) due to the computationally intensive nature of DNN models, while placing the entire model in the cloud brings an extended response delay.

Distributed inference, which involves inference across multiple GPU devices, has emerged as a promising approach to meet the latency requirements of robotic applications and extend the battery lifetime of robots.
This paradigm has been widely adopted in data centers~\cite{xiang2019pipelined, zhuang2023optimizing, hu2022pipeedge}, where numerous GPUs are utilized to speed up large model inference, such as in the case of ChatGPT~\cite{wu2023brief}.
Adopting distributed inference across robots and other powerful GPU devices through the Internet of Things for these robots (robotic IoT) not only accelerates the inference process by leveraging the high computing capabilities of powerful GPUs but also alleviates the local computational burden, thereby reducing energy consumption, making it an ideal solution for robotic applications.

However, all existing parallel methods for distributed inference in the data center are ill-suited for robotic IoT.
In data centers, there are mainly three kinds of parallel methods:
Data parallelism (DP) replicates the model across devices, and lets each replica handle one mini-batch  (i.e., a subset that slices out of an input data set);
Tensor parallelism (TP) splits a single DNN layer over devices; 
Pipeline parallelism (PP) places different layers of a DNN model over devices (layer partitioning) and pipelines the inference to reduce devices’ idling time (pipeline execution).

For DP, the small batch sizes inherent to robotic IoT applications (typically 1) hinder the mini-batch computation, rendering DP inapplicable for robotic IoT.
In the data center, DP is feasible due to the large batch sizes employed (e.g., 16 images), allowing for the division of inputs into mini-batches that still contain several complete inputs (e.g., 2 images). 
However, in robotic IoT, real-time performance is crucial, necessitating immediate inference upon receiving inputs, which typically have smaller batch sizes (e.g., 1 image).
Further splitting these inputs would result in mini-batches containing incomplete inputs (e.g., 1/4 of an image), which cannot be computed in parallel to speed up inference.

TP requires frequent synchronization among devices, leading to unacceptable communication overhead in robotic IoT. 
By partitioning parameter tensors of a layer across GPUs, TP allows concurrent computation on different parts of this tensor but requires an all-reduce communication~\cite{zhuang2023optimizing} to combine computation results from different devices, which entails significant communication overhead. 
Consequently, TP is used mainly for large layers that are too large to fit in one device in data centers and require dedicated high-speed interconnects (e.g., 400 Gbps for NVLink~\cite{li2019evaluating}) even within data centers. 
On the contrary, robots must prioritize seamless mobility and primarily depend on wireless connections, which inherently possess limited bandwidth, as described in Sec.~\ref{sec:bandwidth}, making all-reduce synchronization an unacceptable overhead (e.g., the inference time with TP was up to 143.9X slower than local computation in Sec.~\ref{sec:backgournd-tp}).

\begin{figure}[!t]
    \centering
    \subfloat[Inference Latency\label{fig:robot}]{\includegraphics[width=0.98\linewidth]{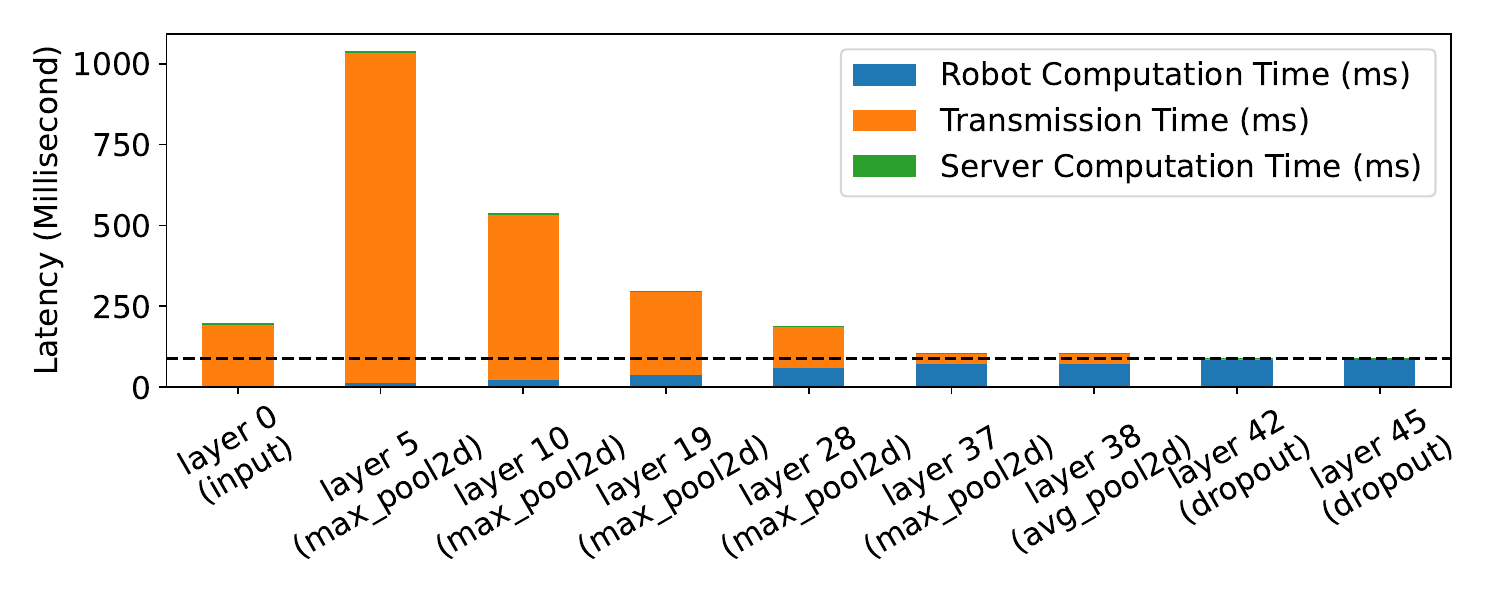}}
    \vfil
    \subfloat[Energy consumption on robot\label{fig:agr}]{\includegraphics[width=0.98\linewidth]{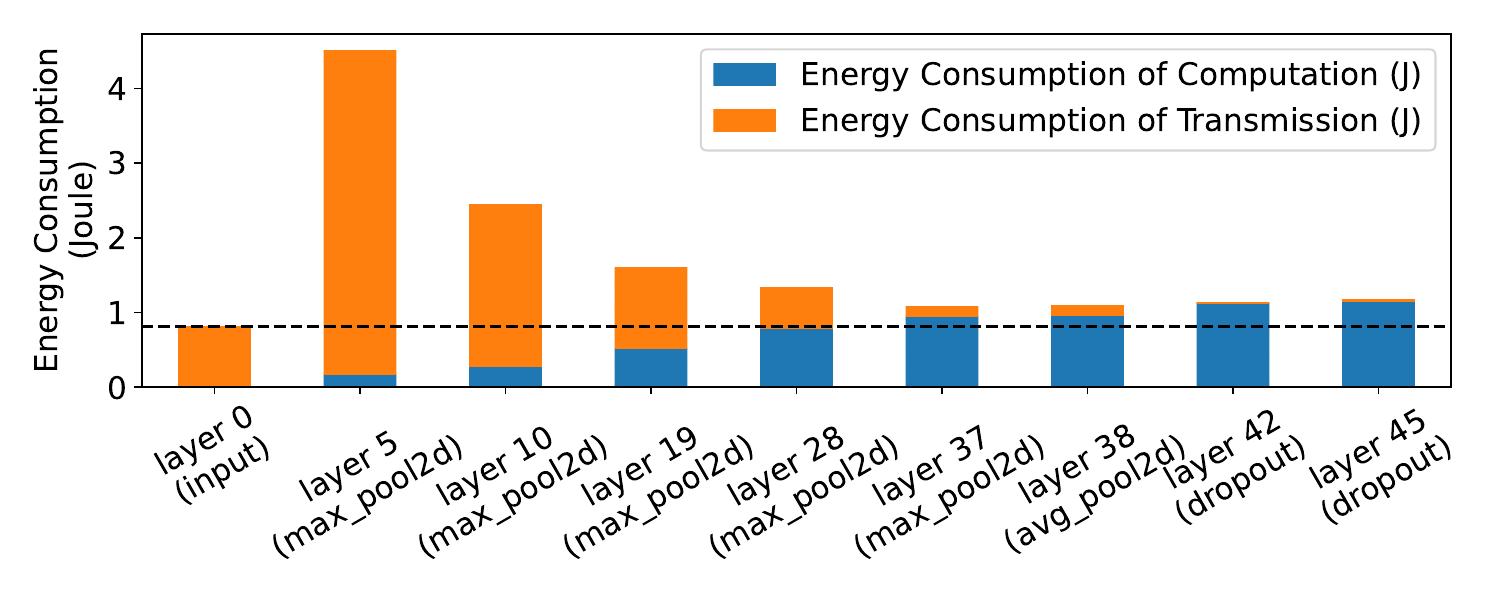}}
    \caption{Existing distributed inference approaches on VGG19~\cite{simonyan2015deep} in our experiments, which adopt PP paradigm with various layer partitioning scheduling strategies.  
    The X-axis of the graph represents different layer partitioning strategies, where `layer i' indicates that all layers up to and including the $i_{th}$ layer are computed on the robot, while the subsequent layers are processed on the GPU server. 
    }
    \label{fig:layer_partitioning}
\end{figure}

Consequently, existing distributed inference approaches~\cite{liang2023dnn, chen2021energy} in robotic IoT are constrained to the PP paradigm. 
Since the PP paradigm in data centers consists of layer partitioning and pipeline execution, where the pipeline execution of PP enhances inference throughput rather than reducing the completion time of a single inference~\cite{crankshaw2020inferline}, the most critical requirement in robotic IoT, existing methods on robotic IoT concentrate on optimizing the layer partitioning aspect of PP to achieve fast and energy-efficient inference.
Based on the fact that the amounts of output data in some intermediate layers of a DNN model are significantly smaller than that of its raw input data~\cite{hu2019dynamic}, DNN layer partitioning strategies constitute various trade-offs between computation and transmission, taking into account application-specific inference speed requirements and energy consumption demands, as shown in Fig.~\ref{fig:layer_partitioning}.

However, existing methods based on the PP paradigm face significant transmission bottlenecks in robotic IoT due to the inherent scheduling mechanism.
The PP paradigm on robotic IoT involves three sequential phases: computing early DNN layers on robots, transmitting intermediate results, and completing inference on a GPU server, where the limited bandwidth of real-world networks often results in transmission time exceeding computation time.
Despite optimal layer partitioning strategies~\cite{liang2023dnn, chen2021energy}, the transmission overhead becomes a substantial bottleneck, accounting for up to 70.45\% of inference time in our experiments, due to the limited bandwidth of robotic IoT. 
This overhead not only slows down inference speed and consumes significantly more energy but also cannot be effectively mitigated by overlapping computation and transmission phases across multiple inference tasks via pipeline execution, which still fails to reduce the completion time of a single inference task~\cite{crankshaw2020inferline}, a crucial aspect for robotic applications.

The key reason for the problem of the above methods is that existing methods conduct layer-granulated scheduling, which divides a single inference task into multiple sequential phases, thereby precluding parallel execution within the scope of an individual inference task.
As transmission time constitutes a substantial portion of the total inference time (approximately half) in existing methods, a novel parallel method that can efficiently overlap computation and transmission within the same inference task has the potential to address this shortcoming, achieving fast inferences.
Note that the robot can not enter low-power sleep mode during the transmission phase due to the need to promptly continue working upon receiving inference results, but can only enter standby mode, when chips like CPU, GPU, and memory consume non-negligible power even when not computing (e.g., 95\% power consumption in our experiments). 
Such a parallel method would reduce the robot's standby time without significantly increasing energy consumption during the computation phase, thereby also decreasing overall energy consumption.

In this paper, we present \xxx{} (Intra-Data Parallel), a high-performance distributed inference system optimized for real-world robotic IoT networks. 
We discovered that operators for each DNN layer (e.g., convolution, ReLU, softmax) can be categorized into two types: local operators and global operators, depending on whether they can be computed independently with partial input. 
For instance, softmax~\cite{liu2016large} requires the complete input vector to calculate the corresponding probability distribution, referring to it as a global operator, while ReLU~\cite{daubechies2022nonlinear} and convolution~\cite{mohammed2020distributed} can be computed with partial input tensor (the elements in the input vector for ReLU and the blocks in the input tensor for convolution), referring to them as local operators.
Since a single local operator like convolution kernel may require multiple calculations per layer, we treat each calculation of the local operator as an independent local operator in this article for easy discussion. 
Local operators are widely used in robotic applications, especially convolution layers in computer vision~\cite{kapao} and point cloud tasks~\cite{agrnav}. 
The local operator granularity provides a finer granularity for \xxx, allowing different local operators of different layers to be computed and transmitted concurrently, enabling the overlap of computation and transmission phases within the same inference task to achieve fast and energy-efficient inference.

The design of \xxx{} is confronted with two major challenges. The first one is how to guarantee the correctness of inference results based on local operator.
We propose Local Operator Parallelism (\xxxparallel), which reduces the granularity of calculation from each layer to each local operator. 
\xxxparallel{} determines the correct input required for different local operators based on their calculation characteristics and processes at first. 
When a part of the local operators in a layer completes the calculation and the tensor composed of these local operators satisfies the input requirements of the local operators in the subsequent layer, the local operators in the subsequent layer can be calculated in advance, without waiting for all local operators of the current layer to be computed in \xxxparallel{}. 
For global operator layers, \xxx{} enforces a synchronization before these layers to combine the complete input for them, as TP's all-reduce communications do. 
In this way, \xxx{} only change the execution sequence of local operators among local operator layers and ensures the calculation correctness of local operator layers through \xxxparallel{} and global operator layers through synchronization.

The second challenge is under \xxxparallel, how to properly schedule the computation and transmission of each local operator to achieve fast and energy-efficient inference under various hardware conditions and network bandwidths. 
\xxx{} places part of the local operator execution on GPU servers and transmits the corresponding part of the input tensor based on \xxxparallel, while computing the rest of the local operators on robot with a novel Local Operator Scheduling Strategy (\xxxschedule). 
\xxxschedule{} formulates the problem of determining which part of the local operators should be executed on robots and which part should be executed on GPU servers as a nonlinear optimization problem (see Sec.~\ref{sec:schedule}), and schedules the computation and transmission of each local operator based on the solution obtained via the differential evolution algorithm~\cite{qin2008differential}.

We implemented \xxx{} in PyTorch~\cite{pytorch} and evaluated \xxx{} on our real-world robots under two typical real-world robotic applications~\cite{kapao,agrnav} and several models common to mobile devices on a larger scale~\cite{sinha2019thin, targ2016resnet, simonyan2015deep, woo2023convnext, xu2022regnet}. 
We compared \xxx{} with two SOTA pipeline parallelism methods as baselines: DSCCS ~\cite{liang2023dnn}, aimed at accelerating inference, and SPSO-GA ~\cite{chen2021energy}, focused on optimizing energy consumption, under different real-world robotic IoT networks environments (namely indoors and outdoors).
Evaluation shows that:
\begin{itemize}
    \item \xxx{} is fast. \xxx{} reduced inference time by 14.9\% \textasciitilde 41.1\% compared to baselines under indoors and outdoors environments.
    \item \xxx{} is energy-efficient. \xxx{} reduced up to 35.3\% energy consumption per inference compared to baselines, due to faster inference speed and limited-increased power consumption against time.
    \item \xxx{} is robust in various robotic IoT environments.
    When the robotic IoT environment changed (from indoors to outdoors), \xxx’s superior performance remained consistent.
    \item \xxx{} is easy to use. It took only three lines of code to apply \xxx{} to existing ML applications.
\end{itemize}

Our main contribution are \xxxparallel, a fine-grained parallel method based on local operators, and \xxxschedule, a new scheduling strategy based on \xxxparallel{} optimized for distributed inference over real-world robotic IoT networks. 
By leveraging these contributions, \xxx{} dramatically reduces the transmission overhead in existing distributed inference on robotic IoT by overlapping the computation and transmission phases within the same inference task, achieving fast and energy-efficient distributed inference on robotic IoT.
We envision that the fast and energy-efficient inference of \xxx{} will foster the deployment of diverse robotic tasks on real-world robots in the field.
\xxx{}'s code is released on \github.

In the rest of this paper, we introduce the background of this paper in Sec.~\ref{sec:background}, give an overview of \xxx{} in Sec.~\ref{sec:overview}, present the detailed design of \xxx{} in Sec.~\ref{sec:design}, evaluate \xxx{} in Sec.~\ref{sec:eva}, and finally conclude in Sec.~\ref{sec:conclusion}.

\section{Background}
\label{sec:background}
\subsection{Characteristics of Robotic IoT}
\label{sec:bandwidth}
In real-world robotic IoT scenarios, devices often navigate and move around for tasks like search and exploration. 
While wireless networks provide high mobility, they also have limited bandwidth. 
For instance, Wi-Fi 6, the most advanced Wi-Fi technology, offers a maximum theoretical bandwidth of 1.2 Gbps for a single stream ~\cite{liu2023first}. 
However, not only the limited hardware resources on the robot can not fully play the potential of Wi-Fi 6~\cite{yang2022mobile}, but also the actual available bandwidth of wireless networks is often reduced in practice due to factors such as movement of the devices~\cite{masiukiewicz2019throughput, pei2013connectivity}, occlusion from by physical barriers~\cite{ding2015performance, sarkar2013effect}, and preemption of the wireless channel by other devices~\cite{adame2021time, ren2018proportional}.

To demonstrate the instability of wireless transmission in real-world situations, we conducted a robot surveillance experiment using four-wheel robots navigating around several given points at 5-40cm/s speed in our lab (indoors) and campus garden (outdoors), with hardware and wireless network settings as described in Sec.~\ref{sec:eva}. 
We believe our setup represents robotic IoT devices' state-of-the-art computation and communication capabilities.  
We saturated the wireless network connection with iperf ~\cite{noauthor_iperf_nodate} and recorded the average bandwidth capacity between these robots every 0.1s for 5 minutes.

\begin{figure}[htp]
    \centering
    \subfloat[Indoors\label{fig:indoors}]{\includegraphics[width=0.48\linewidth]{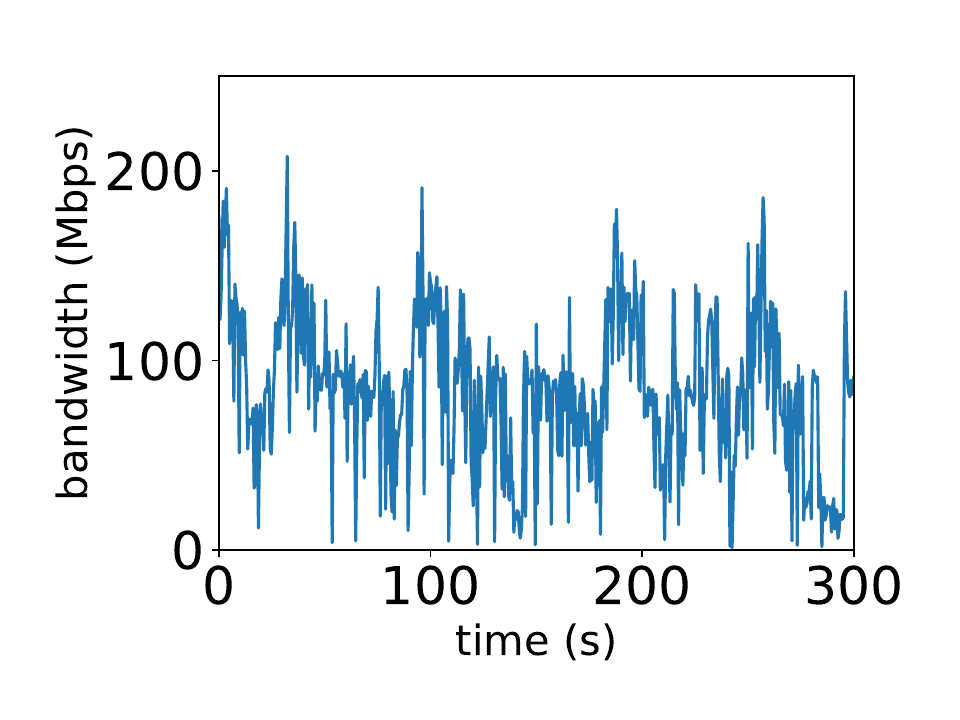}}
    \hfil
    \subfloat[Outdoors\label{fig:outdoors}]{\includegraphics[width=0.48\linewidth]{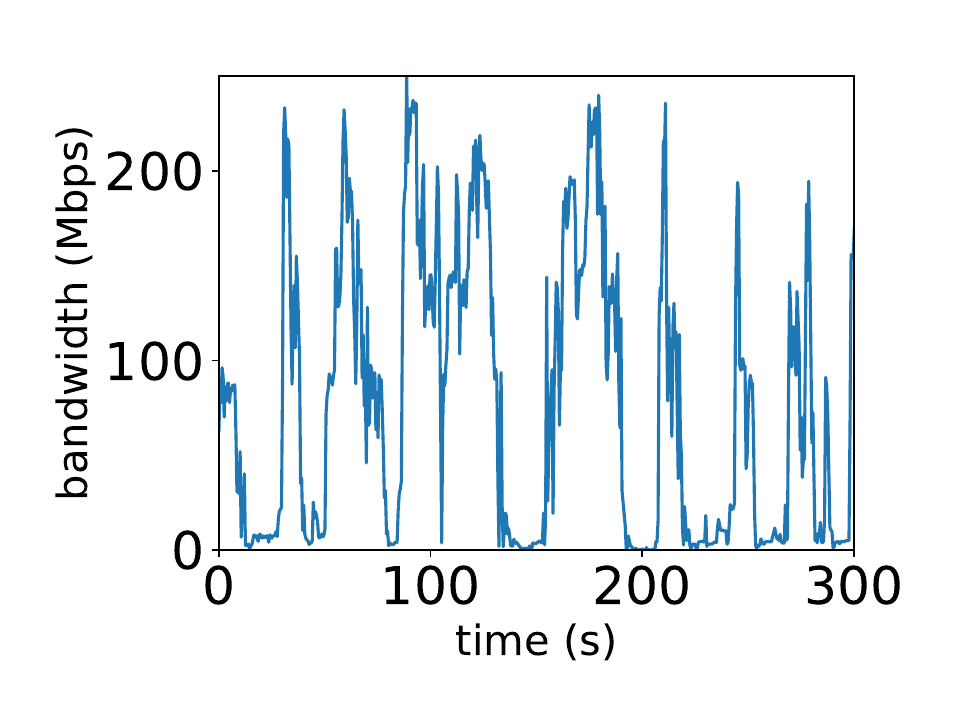}}
    \caption{The instability of wireless transmission between our robot and a base station in robotic IoT networks.}
    \label{fig:bandwidth} 
\end{figure}

The results in Fig.~\ref{fig:bandwidth} show average bandwidth capacities of 93 Mbps and 73 Mbps for indoor and outdoor scenarios, respectively. 
The outdoor environment exhibited higher instability, with bandwidth frequently dropping to extremely low values around 0 Mbps, due to the lack of walls to reflect wireless signals and the presence of obstacles like trees between communicating robots, resulting in fewer received signals compared to indoor environments.

In summary, robotic IoT systems' wireless transmission is constrained by limited bandwidth, both due to the theoretical upper limit of wireless transmission technologies and the practical instability of wireless networks.

\subsection{Characteristics of Data Center Networks} 
Data center networks, which are used for large model inference (e.g., ChatGPT~\cite{wu2023brief}), are wired and typically exhibit higher bandwidth capacity and lower fluctuation compared to robotic IoT networks. 
GPU devices in data centers are interconnected using high-speed networking technologies such as InfiniBand~\cite{wang2011mvapich2} or PCIe~\cite{li2019evaluating}, offering bandwidths ranging from 40 Gbps to 500 Gbps. 
The primary cause of bandwidth fluctuation in these networks is congestion on intermediate switches, which can be mitigated through traffic scheduling techniques implemented on the switches~\cite{noormohammadpour2017datacenter}. 
The stable and high-bandwidth nature of data center networks makes them well-suited for demanding tasks like large model inference, in contrast to the more variable and resource-constrained environments found in robotic IoT networks.

\subsection{Existing distributed inference methods in the data center}

\begin{table*}[t]
    \centering
\begin{tabular}{ccc|c|c|c}
\toprule
 Model(number & Local com- & \multirow[c]{2}{*}{Environment} & Transmission time (s) & Inference time (s) & Percentage(\%) \\
 of parameters & putation time(s) &  & with TP & with TP  & with TP  \\
\midrule
\multirow[c]{2}{*}{MobileNet\_V3\_Small(2M)} & \multirow[c]{2}{*}{0.031($\pm$0.004)} & indoors & 0.698($\pm$0.135) & 1.400($\pm$0.232) & 49.85 \\
 &  & outdoors & 0.901($\pm$0.778) & 1.775($\pm$1.370) & 51.23 \\
\cline{1-6} \cline{2-6}
\multirow[c]{2}{*}{ResNet101(44M)} & \multirow[c]{2}{*}{0.065($\pm$0.005)} & indoors & 7.156($\pm$3.348) & 8.106($\pm$3.403) & 87.95 \\
 &  & outdoors & 8.470($\pm$6.337) & 9.356($\pm$6.328) & 90.46 \\
\cline{1-6} \cline{2-6}
\multirow[c]{2}{*}{VGG19\_BN(143M)} & \multirow[c]{2}{*}{0.063($\pm$0.002)} & indoors & 5.152($\pm$4.873) & 5.444($\pm$4.831) & 70.18 \\
 &  & outdoors & 5.407($\pm$6.673) & 5.759($\pm$6.635) & 93.70 \\
\cline{1-6} \cline{2-6}
\bottomrule
\end{tabular}
    \caption{Average transmission time (Second), inference time (Second), percentage that transmission time accounts for of the total inference time and their standard deviation ($\pm n$) with TP on different models in different environments. ``Local computation'' refers to inference the entire model locally on the robot.}
    \label{tab:tp_time}
\end{table*}

\begin{table*}[t]
    \centering
\begin{tabular}{cc|c|c|c|c}
\toprule
 Model(number& \multirow[c]{2}{*}{Environment} & \multicolumn{2}{|c|}{Power consumption(W)} & \multicolumn{2}{|c|}{Energy consumption(J) per inference} \\
of parameters) &  & Local & TP & Local & TP \\
\midrule
\multirow[c]{2}{*}{MobileNet\_V3\_Small(2M)} & indoors & 6.05($\pm$0.21) & 5.24($\pm$0.19) & 0.3($\pm$0.09) & 7.33($\pm$1.21) \\
 & outdoors & 6.05($\pm$0.21) & 5.11($\pm$0.28) & 0.3($\pm$0.09) & 9.08($\pm$7.0) \\
\cline{1-6}
\multirow[c]{2}{*}{ResNet101(44M)} & indoors & 11.27($\pm$0.51) & 4.97($\pm$0.16) & 0.93($\pm$0.19) & 40.28($\pm$16.91) \\
 & outdoors & 11.27($\pm$0.51) & 4.9($\pm$0.23) & 0.93($\pm$0.19) & 45.8($\pm$30.98) \\
\cline{1-6}
\multirow[c]{2}{*}{VGG19\_BN(143M)} & indoors & 14.86($\pm$0.43) & 4.88($\pm$0.29) & 1.19($\pm$0.18) & 26.55($\pm$23.56) \\
 & outdoors & 14.86($\pm$0.43) & 4.87($\pm$0.27) & 1.19($\pm$0.18) & 28.06($\pm$32.33) \\
\cline{1-6}
\bottomrule
\end{tabular}
    \caption{Power consumption against time (Watt) and energy consumption per inference (Joule) with standard deviation ($\pm n$) with TP on different models in different environments. ``Local'' represents ``Local computation''}
    \label{tab:tp_power}
\end{table*}

\myparagraph{Data parallelism} 
DP~\cite{xiang2019pipelined} is a widely used technique in distributed inference that splits input data across multiple GPU devices to perform parallel inference. 
Each device has a complete copy of the model and processes a portion of the input data independently, combining the results to produce the final output. 
This approach improves throughput by distributing the workload across devices. 
However, data parallelism's scalability is limited by the total batch size~\cite{narayanan2021efficient}, which is especially challenging in robotic IoT applications. 
In these applications, smaller batch sizes are common because of the need for quick responses to the environment and immediate inference when inputs are received. 
For example, in our experiments, the robot continuously receives the latest images from the camera for inference, with a batch size of only 1, which cannot be further divided into mini-batches, a crucial requirement for effective data parallelism.

\myparagraph{Tensor parallelism} 
\label{sec:backgournd-tp}
TP~\cite{zhuang2023optimizing} is a distributed inference technique that splits a model's layer parameters across multiple devices, each storing and computing a portion of the parameter tensors. 
This approach requires an all-reduce communication step after each layer to combine results from different devices, which introduces significant overhead, especially for large DNN layers. 
To mitigate this, TP is typically used across GPUs within the same server in data centers, using fast intra-server GPU-to-GPU links like NVLink ~\cite{li2019evaluating}, which is helpful when the model is too large for a single device. 
However, in robotic IoT, the limited bandwidth makes the communication cost of TP prohibitively high. 
Our evaluation on the same testbed as Sec.~\ref{sec:eva} of DINA~\cite{mohammed2020distributed}, a state-of-the-art TP method,  shows that transmission time takes up 49\% to 94\% of the total inference time due to all-reduce communication for each layer, making TP's inference time 45.2X to 143.9X longer than local computation in Table~\ref{tab:tp_time}. 
Although TP has lower power consumption (13.4\% to 67.3\% less than local computation), the extended transmission times significantly increase energy consumption per inference by 28.5X to 62.7X in Table~\ref{tab:tp_power}. 
As TP greatly extends inference time, making it impractical for real-world robotic applications, we did not consider it further in this paper.

\myparagraph{Pipeline parallelism}
PP~\cite{hu2022pipeedge} is a distributed inference technique that partitions DNN model layers across multiple devices(layer partitioning), forming an inference pipeline for concurrent processing of multiple tasks. 
While PP can increase throughput and resource utilization via pipeline execution, it primarily focuses on enhancing overall throughput rather than reducing single-inference latency~\cite{crankshaw2020inferline}, which is crucial in robotic IoT. 
As a result, existing distributed inference approaches~\cite{liang2023dnn, chen2021energy} in robotic IoT primarily adopt PP paradigm and focus on layer partitioning to achieve fast and energy-efficient inference, with two main categories based on their optimization goals: accelerating inference for diverse DNN structures~\cite{hu2019dynamic, mohammed2020distributed, kang2017neurosurgeon, liang2023dnn, xue2021ddpqn} and optimizing robot energy consumption during inference~\cite{wu2019efficient, lin2019cost, chen2021energy}. 
However, both kinds of methods suffer from the transmission bottleneck inherent to PP's scheduling mechanism, which can be eliminated by \xxx.

\subsection{Other methods to speed up DNN Models Inference on Robotic IoT}
\myparagraph{Compressed communication}
Compressed communication is essential for efficient distributed inference in wireless networks, as it significantly reduces communication overhead through techniques such as quantization and model distillation. 
Quantization~\cite{defossez2021differentiable,gheorghe2021model,gong2020vecq} reduces the numerical precision of model weights and activations, minimizing the memory footprint and computational requirements of deep learning models by converting high-precision floating-point values (e.g., 32-bit) to lower-precision representations (e.g., 8-bit) with minimal loss of model accuracy. 
Model distillation~\cite{gou2021knowledge,lin2020ensemble,wang2021knowledge}, on the other hand, involves training a smaller, more efficient "student" model to mimic the behavior of a larger, more accurate "teacher" model by minimizing the difference between their outputs, allowing the distilled student model to retain much of the teacher model's accuracy while requiring significantly fewer resources. 
These model compression methods complement distributed inference by achieving faster inference speed through model modifications, potentially sacrificing some accuracy with smaller models, while distributed inference realizes fast inference without loss of accuracy by intelligently scheduling computation tasks across multiple devices.

\myparagraph{Inference Job scheduling}
Significant research efforts have been devoted to exploring inference parallelism and unleashing the potential of layer partition to accelerate DNN inference, such as inference job scheduling, which aims to accelerate multiple DNN inference tasks by optimizing their execution on various devices under different network bandwidths while considering application-specific inference speed requirements and energy consumption demands. 
For instance, ~\cite{altamimi2015energy, elgazzar2014cloud} support online scheduling of offloading inference tasks based on the current network and resource status of mobile systems while meeting user-defined energy constraints, while ~\cite{fang2017qos} focus on optimizing DNN inference workloads in cloud computing using a deep reinforcement learning based scheduler for QoS-aware scheduling of heterogeneous servers, aiming to maximize inference accuracy and minimize response delay. 
However, while these methods focus on overall optimization in multi-task scenarios involving multi-robots, they do not address the optimization of single inference tasks and are thus orthogonal to distributed inference for a single inference, where improved distributed inference can provide faster and more energy-efficient inference for these scenarios.

\section{Overview}
\label{sec:overview}
\subsection{Workflow of \xxx}
\begin{figure}[htp]
    \centering
    \includegraphics[width=0.98\linewidth]{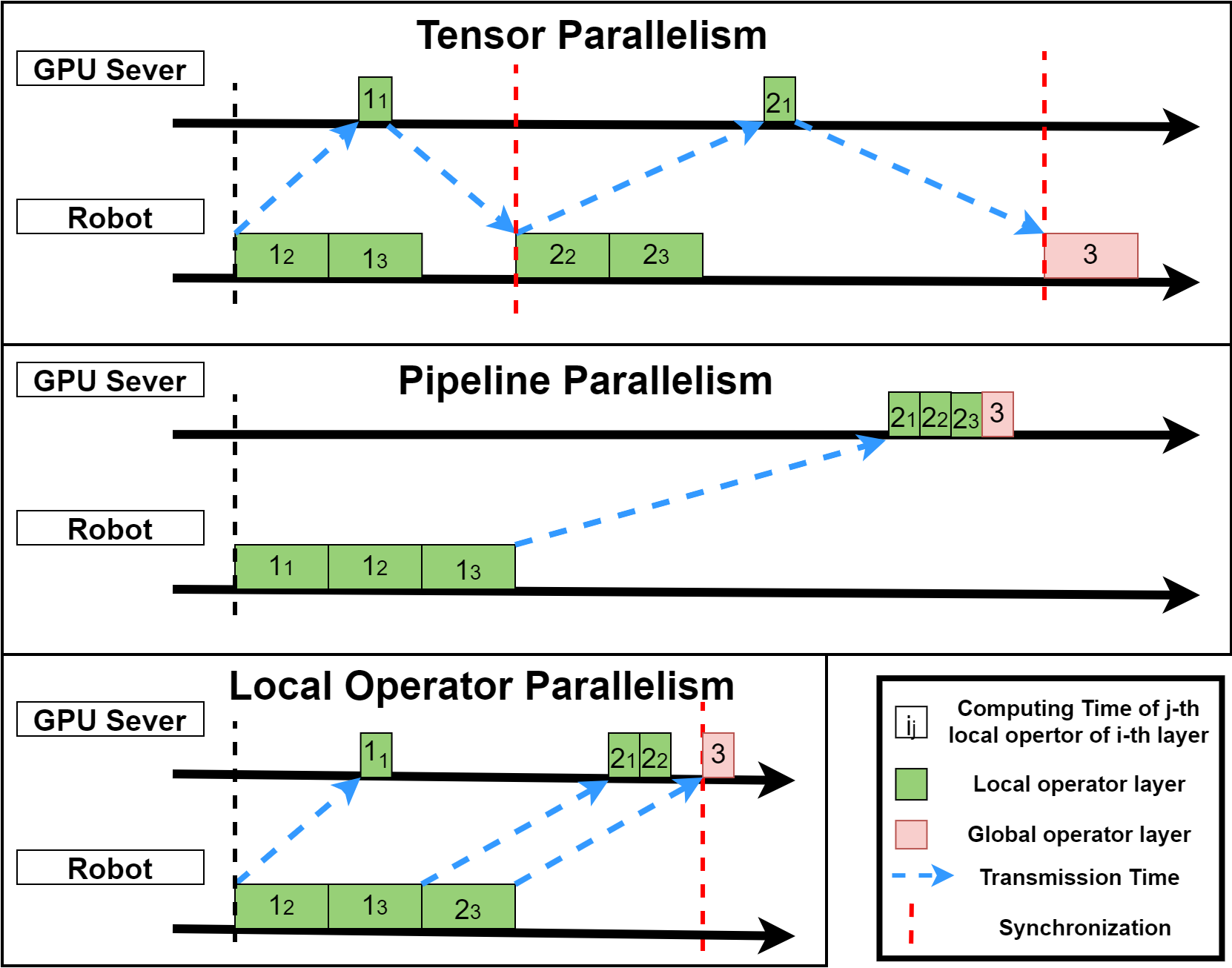}
    \caption{Workflow of \xxx. 
    Each local operator layer have to complete the calculation of three local operators, and the same local operator in the three cases has the same computation time on robots and GPU servers, as well as the corresponding transmission time. 
    The output tensor volume of layer 2 is larger than that of layer 1, resulting in longer transmission times for local operators in layer 2, and PP selects a layer partition strategy at layer 1~\cite{liang2023dnn}.
    }
    \label{fig:overview}
\end{figure}

Fig.~\ref{fig:overview} presents the workflow of \xxx{} and compares it with TP and PP under robotic IoT networks with limited bandwidth, illustrating why existing methods suffer from transmission overhead and how \xxx{} solves this issue via its \xxxparallel.

While TP can place some local operator execution on the GPU server, it requires an all-reduce communication~\cite{zhuang2023optimizing} to combine computation results from different devices, which entails significant communication overhead (as shown by the red dotted lines for synchronization Fig.~\ref{fig:overview}). 
Although the layer partition algorithm~\cite{liang2023dnn} can be used to minimize overall inference time, the transmission time still becomes a significant bottleneck, as illustrated by the extremely long transmission in Fig.~\ref{fig:overview}.

To alleviate the transmission overhead in distributed inference, \xxx{} overlaps the computation and transmission of different local operators from different local operator layers, as shown in Fig.~\ref{fig:overview}. 
Compared with TP, \xxx{} cancels the synchronization of the all-reduce communication for local operator layers and ensures that each local operator can get the required input in time and obtain the correct calculation results through \xxxparallel, rather than relying on all-reduce communication for local operator layers. 
\xxx{} maintains synchronization for global operator layers, which do not have local operators and require the complete input (the entire output tensor of the previous layer) to perform the calculation, ensuring the correctness of the global operator layers' inference. 
Compared with PP, \xxx{} starts transmitting some local operators and the corresponding input in advance, without waiting for all local operators in the current local operator layer to complete the calculation.
In this way, \xxx{} achieves much faster inference compared with existing distributed inference methods.

Moreover, the idle time on the robot (when the robot is not computing, as shown in Fig.~\ref{fig:overview}) consumes significant energy.
This is because the robot cannot enter a low-power sleep mode while waiting for the final inference result from the GPU server, as it has to promptly continue working when it receives the inference results. 
During the standby phase (idle time), chips like CPU, GPU, and memory consume non-negligible power even when not computing, due to the static power consumption rooted in transistors' leakage current~\cite{kim2003leakage}. 
Meanwhile, we found that wireless network cards consume only 0.21 Watt for transmission during the idle time, while the robot consumes 13.35 Watt during computing. 
In this way, \xxx{} dramatically reduces the idle time on the robot, alleviating the energy wasted by standby mode, and increases a negligible amount of network card transmission power consumption during computing, thereby reducing the overall energy consumption for each inference.

To achieve the workflow shown in Fig.~\ref{fig:overview}, the design of \xxx{} must tackle two problems: guaranteeing the correctness of inference results based on local operators and scheduling the computation and transmission of each local operator. 
In Sec.~\ref{sec:parallel}, we will explain how \xxx{} ensures that each local operator can still obtain the correct calculation result via \xxxparallel, and in Sec.~\ref{sec:schedule}, we will discuss how \xxx{} achieves fast and energy-efficient inference through its \xxxschedule.

\subsection{Architecture of \xxx}
\label{sec:architecture}
\begin{figure}[htp]
    \centering
    \includegraphics[width=0.98\linewidth]{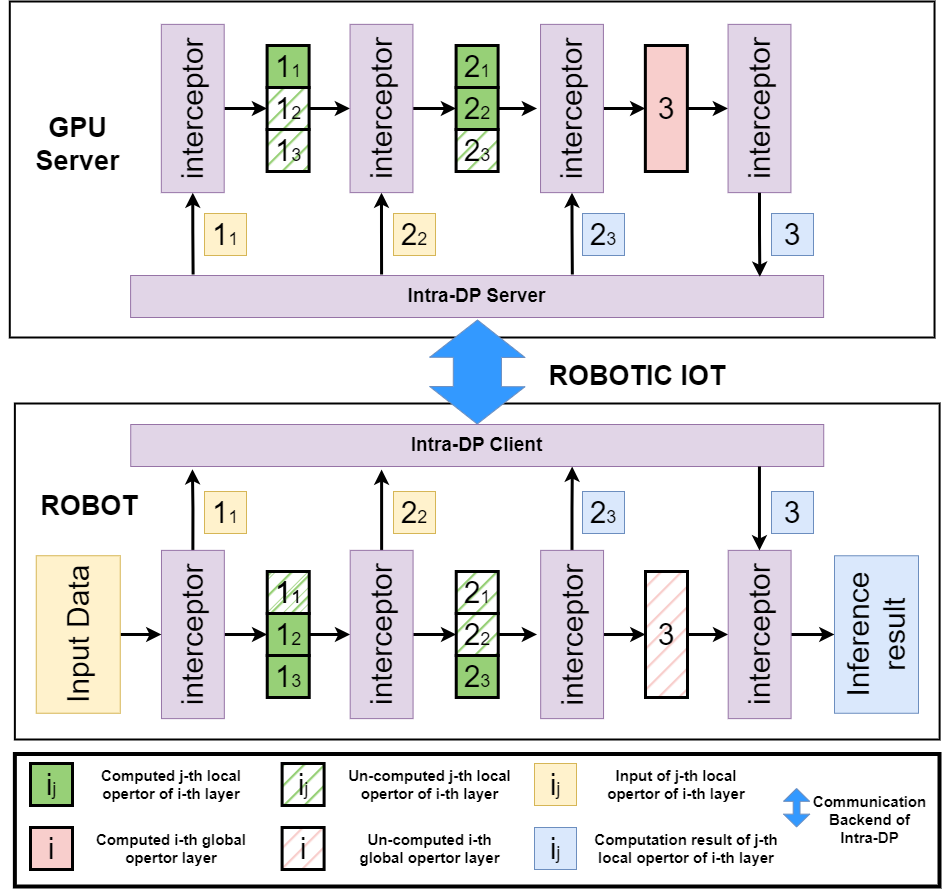}
    \caption{Architecture of \xxx. 
    The core components of \xxx{} are highlighted in purple.
    \xxx{} adopts the same scheduling scheme as in Fig.~\ref{fig:overview}.}
    \label{fig:architecture}
\end{figure}

Fig.~\ref{fig:architecture} shows the architecture of \xxx, which adds an interceptor for each DNN layer to flexibly split the input tensor and combine the output tensor for each operator. 
Compared with the original model inference process on the robot, \xxx{} only increases the time cost of interceptors, which is the time cost of splitting the input tensor and combining the output tensor. 
The time cost of splitting the input tensor is negligible because the data transfer can be completed through the backend processes of the \xxx{} client and server while the local operators assigned to be executed on the robot and GPU server continue to perform subsequent layer calculations. 
The time cost of combining the output tensor is mainly bound by the time when the device on the other side completes the corresponding computation and transmission, causing prolonged waiting time. 
\xxx{} formulates such waiting time into the nonlinear optimization problem in its \xxxschedule, minimizing the waiting time and implementing scheduling schemes on local operators with a higher degree of parallelism. 
In this way, \xxx{} only increases negligible extra time on system cost and achieves faster inference via \xxxparallel{} and \xxxschedule.

To address frequent fluctuations in real-world wireless networks of robotic IoT, \xxx{} generates optimal local operator schedule plans for the DNN model under different bandwidth conditions in advance. 
During inference, \xxx{} predicts the network bandwidth using mature tools~\cite{yue2017linkforecast} in the field of wireless transmission and adopts the corresponding schedule plan based on the predicted bandwidth. 
To ensure that \xxx{} can flexibly switch among various schedule plans, it keeps a copy of the model on the robot at the GPU server (Fig.~\ref{fig:architecture}), avoiding unnecessary transmission when migrating the parameters of local operators between robots and the GPU server. 
It is important to note that the model inference time, typically tens or hundreds of milliseconds, is finer (smaller) than the granularity (or frequency) of bandwidth fluctuation in real-world robotic IoT networks, as shown in Fig.~\ref{fig:bandwidth}.
Therefore, we assume that the network bandwidth of robotic IoT during each inference is stable, while the network bandwidth for different inferences may differ.

\section{Detailed Design}
\label{sec:design}
\subsection{Local Operator Parallelism}
\label{sec:parallel}
\xxxparallel{} guarantees the correctness of inference results by determining the correct input required for different local operators based on their calculation characteristics and processes. 
We summarize three classes of local operators common in models used on mobile devices: 

\begin{itemize}
    \item Element-wise local operator. 
    This class of operators compute each element of the input tensor separately, requiring only the corresponding element from the input tensor to perform the calculation. 
    They are widely used in activation functions such as ReLU~\cite{daubechies2022nonlinear}, Sigmoid~\cite{yin2003flexible}, SiLU~\cite{jocher2021ultralytics}. 
    However, it is important to note that some activation functions, like softmax~\cite{liu2016large}, require all elements for computation and are not considered local operators, but global operators.

    \item Block-wise local operator.
    This class of operators require a block at the corresponding position in the input tensor and are widely used in layers associated with convolution, such as convolution~\cite{mohammed2020distributed}, maxpool~\cite{sun2021ampnet}. 
    The size of the input blocks is determined by the parameters set by the corresponding layer~\cite{conv2d}, including the size of the convolution kernel, padding, and dilation.

    \item Row-wise local operator.
     This class of operators requires rows of the input tensor and are widely used in layers associated with matrix operations, such as addition~\cite{zee1996law} and multiplication~\cite{fatahalian2004understanding}.
     The rows required for computation, ensuring that the correct input is obtained for each local operator to perform its respective calculation, are determined by the matrix calculation principles as following:
\begin{equation*}
\left(
\begin{smallmatrix}
a_{1} \\
\vdots \\
a_{m}
\end{smallmatrix}
\right)
\times
\left(
\begin{smallmatrix}
b_{1} & \cdots & b_{n}
\end{smallmatrix}
\right)
=
\left(
\begin{smallmatrix}
c_{11} & \cdots & c_{1n} \\
\vdots & \ddots & \vdots \\
c_{m1} & \cdots & c_{mn}
\end{smallmatrix}
\right)
\end{equation*}
Row-wised local operators split the input matrix and keep a copy of the layer parameter matrix on different devices, reducing transmission volume and avoiding the synchronization needed to combine calculation results from different devices. 
This is in contrast to TP, which splits the layer parameter matrix and transfers a copy of the input matrix to different devices. 
The calculation result of row $a_{1}$ is $\left( c_{11} \cdots c_{1n}\right)$, which is also a row and can be directly computed by the next matrix operation layer. 
And \xxxparallel{} treats matrices with only one row as global operators.
\end{itemize}

After obtaining the required input for each local operator and the corresponding input position in the previous layer through the above analysis, \xxxparallel{} determines which local operators need to be computed in the previous layer to obtain the input for the current local operator. 
This establishes the dependency between local operators, which should be considered when scheduling local operators in \xxxschedule. 
Notice that when the result of an operator is used by several local operators in the following layer, especially for block-wise local operators, \xxxparallel{} allows some operators to be repeatedly computed by the robot and the GPU server, which avoids synchronization with high transmission costs in robotic IoT by introducing a small amount of redundant calculation. 
And we leave the support for additional types of local operators as future work.

\subsection{Local Operator Scheduling Strategy}
\label{sec:schedule}
\xxxschedule{} formulates the problem of scheduling local operators as a nonlinear optimization problem, modeled as follows:

First, we define $OP_{i}$ as the set of operators in the $i_{th}$ layer, including both local and global operators. 
When the $i_{th}$ layer is a global operator layer, $\left| OP_{i} \right|$ is 1, as it only has one operator. 
We then define $X_{i} \subseteq OP_{i}$ as the set of operators executed on robots and $Y_{i} \subseteq OP_{i}$ as the set of operators executed on the GPU server, where $X_{i} \cup Y_{i} = OP_{i}$.
$X_{i} \cap Y_{i} \neq \emptyset$ when the result of an operator is used by several local operators in the following layer, especially for block-wise local operators; otherwise, $X_{i} \cap Y_{i} = \emptyset$.

Next, we denote the completion time of the $i_{th}$ layer on robots as $T_{robot}^{i}$ and that on the GPU server as $T_{server}^{i}$, as shown in Fig.~\ref{fig:overview}. 
We define $compute\left( X \right)$ as the estimated computation time of $X$ and $transmit\left( X \right)$ as the estimated transmission time of $X$ under the given bandwidth, leading to the following formula:
$$
T_{robot}^{i} = 
\begin{cases}
\begin{aligned}
&compute\left( X_{0} \right)     & i=0 \\
&T_{robot}^{i-1} + compute\left( X_{i} \right)   & i > 0, M_{i} = \emptyset\\
&MAX\left(T_{robot}^{i-1}, T_{server}^{i-1} + \right.\\
&\left. transmit\left( M_{i} \right) \right) + compute\left( X_{i} \right)   & i >0, M_{i} \neq \emptyset\\
\end{aligned}
\end{cases}
$$
$$
T_{server}^{i} = 
\begin{cases}
\begin{aligned}
&transmit\left( Y_{0} \right) +compute\left( Y_{0} \right)     & i=0 \\
&T_{server}^{i-1} + compute\left( Y_{i} \right)   & i > 0, N_{i} = \emptyset\\
&MAX\left(T_{server}^{i-1}, T_{robot}^{i-1} + \right.\\
&\left. transmit\left( N_{i} \right) \right) + compute\left( Y_{i} \right)   & i >0, N_{i} \neq \emptyset\\
\end{aligned}
\end{cases}
$$
Here, $M_{i} = parent\left( X_{i} \right) - X_{i-1}$ and $N_{i} = parent\left( Y_{i} \right) - Y_{i-1}$, where $parent\left(X_{i}\right)$ is the set of operators found by \xxxparallel{} in the ${i-1}_{th}$ layer, whose outputs form the inputs of $X_{i}$. 
The $MAX$ function is used to minimize the idle time when combining the input tensor of this layer, and $transmit\left(X\right)$ includes not only its own transmission time but also the wait time for the previous transmission to complete.

Next, we present the corresponding objective function and constraints of a DNN model with N layers as a nonlinear optimization problem:
\begin{alignat}{2}
\min \quad& T_{robot}^{N} & \tag{1}\\
\mbox{s.t.}\quad
&M_{i} = \emptyset,\forall i \in \Pi &\tag{2} \\
&N_{i} = \emptyset,\forall i \in \Pi &\tag{3}
\end{alignat}
Here, the layers in $\Pi$ are those whose output data amounts are larger than the raw input data. 
Constraints are inspired by the key observation used in existing layer partitioning methods to limit the transmission overhead, which states that the output data amounts in some intermediate layers of a DNN model are significantly smaller than that of its raw input data~\cite{hu2019dynamic}.

\xxxschedule{} solves the above nonlinear optimization problem using the differential evolution algorithm~\cite{qin2008differential} and schedules the computation and transmission of each local operator based on the obtained solution.
It is important to note that when applying \xxx{} to a special model without any local operator layers, \xxxschedule{} will degrade to the existing layer partitioning method. When applying \xxx{} to DNN models with complex structures as directed acyclic graphs (DAGs) (e.g., MobileNet~\cite{sinha2019thin}, ResNet~\cite{targ2016resnet}), rather than simple chain-like DNN models (e.g., VGG19~\cite{simonyan2015deep}) as above, the ${i-1}_{th}$ layer in the above modeling process should be replaced by the parent layer in the corresponding DAG.

\subsection{Algorithms of \xxx}

\begin{algorithm}[htbp]
\caption{\small \xxx\_client\label{alg:client}}
\SetKwInput{KwInput}{Input}                
\SetKwInput{KwOutput}{Output}              
\DontPrintSemicolon
  \KwInput{\small Data input for inference $input$; DNN model $model$}
  \KwOutput{\small The inference result $ret$ }
  \KwData{\small input of $i_{th}$ layer $Z_{i}$; schedule plan of $i_{th}$ layer under the $b$ bandwidth $X_{i}^{b}$, $M_{i}^{b}$,$N_{i}^{b}$}
   \tcp{\footnotesize profile phase on robot}

   $info\_robot = ProfileModel(model)$

   $SendToServer(model,info\_robot)$

   $X, M, N = ReceiveFromServer()$

   \tcp{\footnotesize inference phase on robot}

   $b = PredictsBandwidth()$

   $Z_{0} = input$

   \ForEach {$i_{th} \quad layer \quad in\quad model$ }{

    \If{$M_{i}^{b} \neq \emptyset$}
    {

    $Z_{i} = combine(Z_{i}, ReceiveFromServer())$
    
    }

    \If{$N_{i}^{b} \neq \emptyset$}
    {

    $SendToServer(Z_{i}, N_{i}^{b})$
    
    }

    \If{$X_{i}^{b} \neq \emptyset$ and $Z_{i} \neq \emptyset$}
    {

     $Z_{i+1} = compute(Z_{i},X_{i}^{b})$
    
    }
    \Else{

    $Z_{i+1} = \emptyset$
    
    }

  } 
  
  $ret = Z_{N+1}$
  
\KwRet{$ret$}
\end{algorithm}

\begin{algorithm}[htbp]
\caption{\small \xxx\_server\label{alg:server}}
\SetKwInput{KwInput}{Input}                
\SetKwInput{KwOutput}{Output}              
\DontPrintSemicolon
  \KwData{\small input of $i_{th}$ layer $Z_{i}$; schedule plan of $i_{th}$ layer under the $b$ bandwidth $Y_{i}^{b}$, $M_{i}^{b}$,$N_{i}^{b}$}
   \tcp{\footnotesize profile phase on server}

   $model,info\_robot = ReceiveFromClient()$

   $info\_server = ProfileModel(model)$

   $X, Y, M, N = LOSS(info\_robot,info\_server)$

   $SendToClient(X,M,N)$

   \tcp{\footnotesize inference phase on robot}

   $b = PredictsBandwidth()$

   $Z_{0} = \emptyset$

   \ForEach {$i_{th} \quad layer \quad in\quad model$ }{

    \If{$M_{i}^{b} \neq \emptyset$}
    {

    $SendToClient(Z_{i}, M_{i}^{b})$
    
    }

    \If{$N_{i}^{b} \neq \emptyset$}
    {  

    $Z_{i} = combine(Z_{i}, ReceiveFromClient())$
    
    }

     \If{$Y_{i}^{b} \neq \emptyset$ and $Z_{i} \neq \emptyset$}
    {

     $Z_{i+1} = compute(Z_{i},Y_{i}^{b})$
    
    }
    \Else{

    $Z_{i+1} = \emptyset$
    }

  } 
  
\end{algorithm}

Here, we present the algorithm of \xxx{} for both the client side on the robot and the server side on the GPU server, as shown in Fig.~\ref{fig:architecture}. 
The client part is given in Alg.~\ref{alg:client} and the server part is given in Alg.~\ref{alg:server}. 
Both sides must first enter the profile phase and provide the basic information for \xxxschedule{} (including the model structure with local and global operators, and the estimate functions $compute$ and $transmit$), represented in $info\_robot$ and $info\_server$. 
Then, \xxx{} generates the schedule plan of local operators under various bandwidths. Compared with the inference time, the solution of \xxxschedule{} takes longer to complete, but these profiles are completed in advance, and only need to select the corresponding schedule plan according to the actual bandwidth during actual use. 
The copy of the model on the GPU server (Fig.~\ref{fig:architecture}, line 1 in Alg.~\ref{alg:server}) allows \xxx{} to switch among various schedule plans flexibly.

\section{Implementation}
\label{sec:implement}
\begin{figure}[htp]
    \centering
    \includegraphics[width=0.8\linewidth]{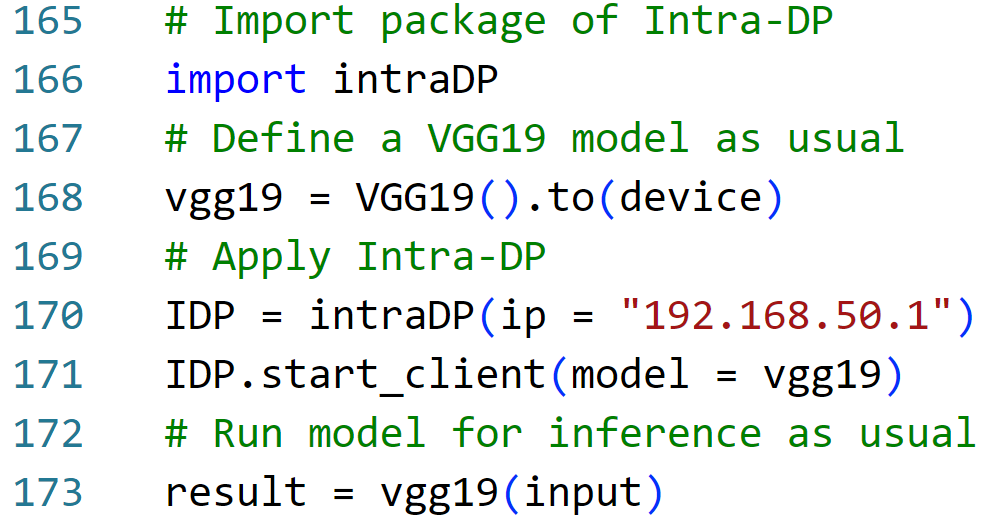}
    \caption{An example of applying \xxx{} to a VGG19~\cite{simonyan2015deep} model, where ``192.168.50.1'' is the IP address of the GPU server.}
    \label{fig:example}
\end{figure}
We implement \xxx{} on Python and PyTorch. \xxx{} is easy to use and requires only three lines of code to apply to existing ML applications, as shown in Fig.~\ref{fig:example}.
This is achieved by hooking around the forward method of the model, and in the first forward call we profile the model using the default PyTorch profiler and schedule; then we intercept and parallelize all the following forward calls as scheduled.

\section{Evaluation}
\label{sec:eva}
\begin{figure}[!t]
    \centering
    \subfloat[Four-wheeled robot\label{fig:robot}]{\includegraphics[width=0.48\linewidth]{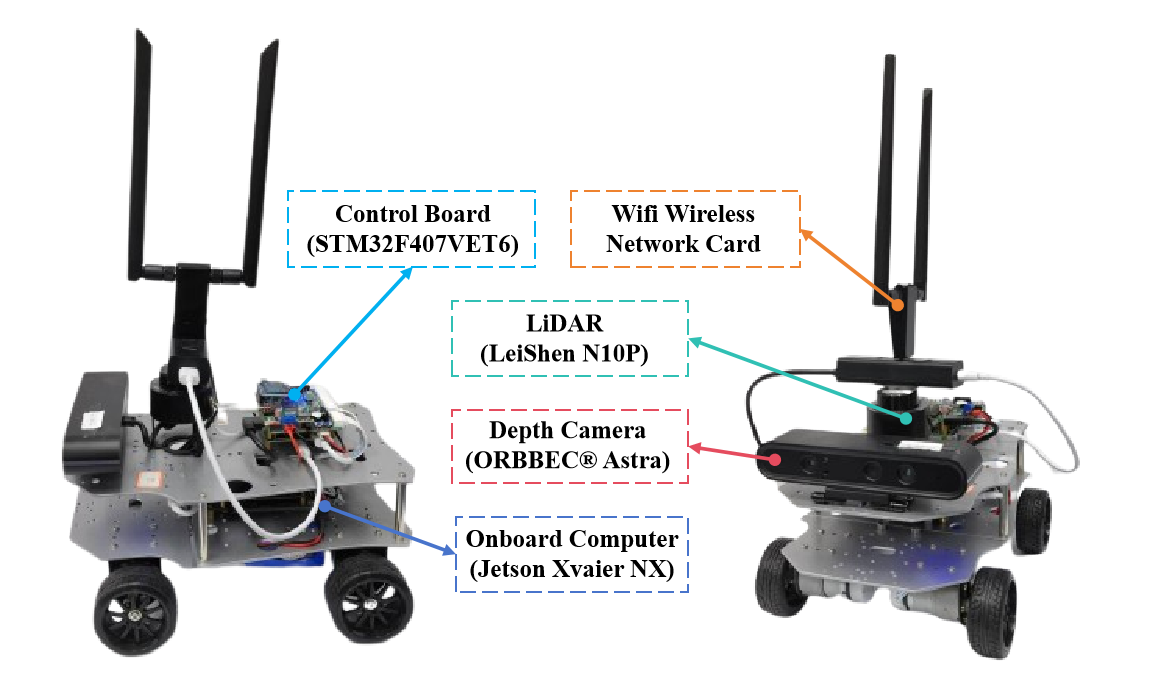}}
    \hfil
    \subfloat[Air-ground robot\label{fig:agr}]{\includegraphics[width=0.48\linewidth]{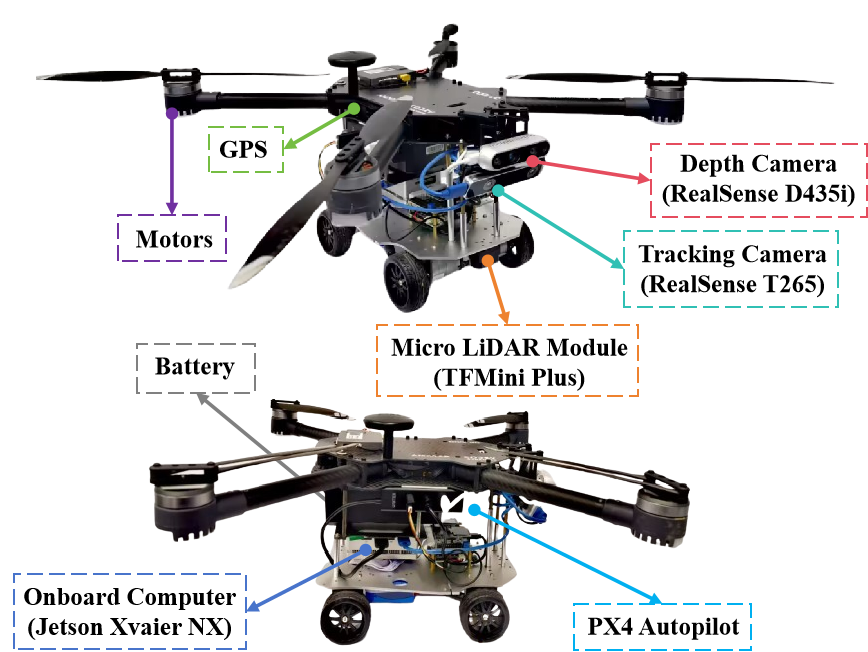}}
    \caption{The detailed composition of the robot platforms}
\end{figure}

\begin{table}[!t]
\centering
\begin{tabular}{|c|c|c|c|}
\hline
        & inference & transmission & standby \\ \hline
Power (W) &     13.35        &       4.25        &    4.04   \\ \hline
\end{tabular}
\caption{Power consumption (Watt) of our robot in different states.}
\label{tab:energydefault}
\end{table}

\myparagraph{Testbed}
The evaluation was conducted on a custom four-wheeled robot (Fig~\ref{fig:robot}), and a custom air-ground robot(Fig~\ref{fig:agr}).
They are equipped with a Jetson Xavier NX~\cite{jetsonnx} 8G onboard computer that is capable of AI model inference with local computation resources. 
The system runs Ubuntu 20.04 with ROS Noetic and a dual-band USB network card (MediaTek MT76x2U) for wireless connectivity. 
The Jetson Xavier NX interfaces with a Leishen N10P LiDAR, ORBBEC Astra depth camera, and an STM32F407VET6 controller via USB serial ports. 
Both LiDAR and depth cameras facilitate environmental perception, enabling autonomous navigation, obstacle avoidance, and SLAM mapping. 
The GPU server is a PC equipped with an Intel(R) i5 12400f CPU @ 4.40GHz and an NVIDIA GeForce GTX 2080 Ti 11GB GPU, connected to our robot via Wi-Fi 6 over 80MHz channel at 5GHz frequency in our experiments.

Tab.~\ref{tab:energydefault} presents the overall on-board energy consumption (excluding motor energy consumption for robot movement) of the robot in various states: inference (model inference with full GPU utilization, including CPU and GPU energy consumption), transmission (communication with the GPU server, including wireless network card energy consumption), and standby (robot has no tasks to execute).
Notice that different models, due to varying numbers of parameters, exhibit distinct GPU utilization rates and power consumption during inference.

\begin{figure}[!t]
    \centering
    \subfloat[Targeted people]{\includegraphics[width=0.48\linewidth]{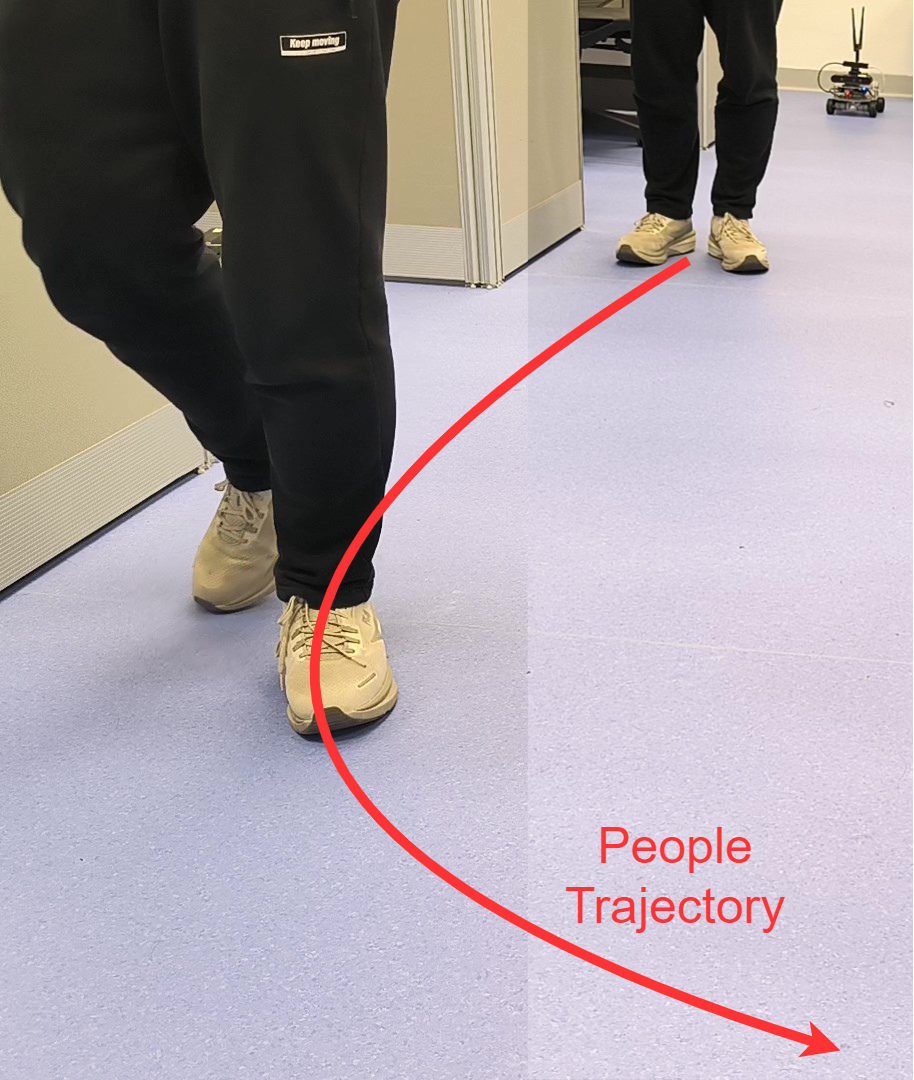}}
    \hfil
    \subfloat[Robot moving trajectory]{\includegraphics[width=0.48\linewidth]{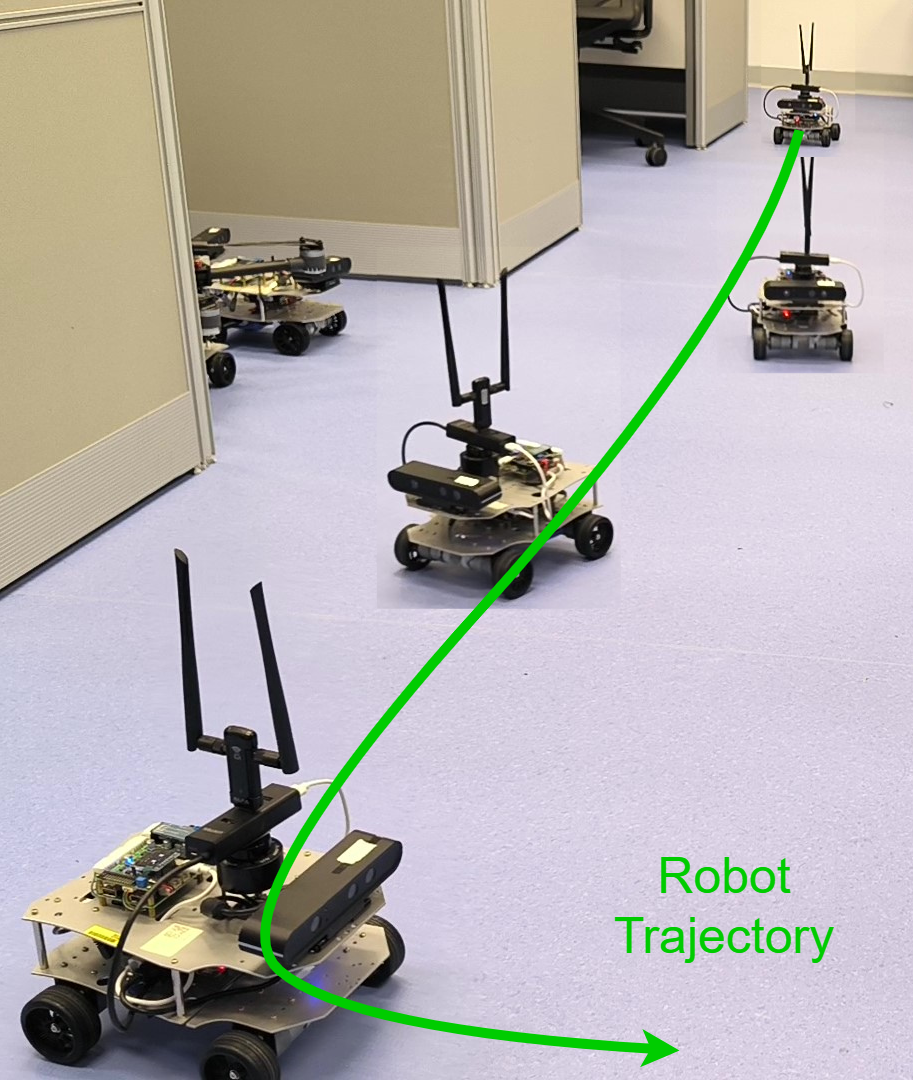}}
    \caption{A real-time people-tracking robotic application on our robot based on a well-known human pose estimation ML model, Kapao~\cite{kapao}.}
    \label{fig:kapao}
\end{figure}

\begin{figure}[!t]
    \centering
    \includegraphics[width=0.98\linewidth]{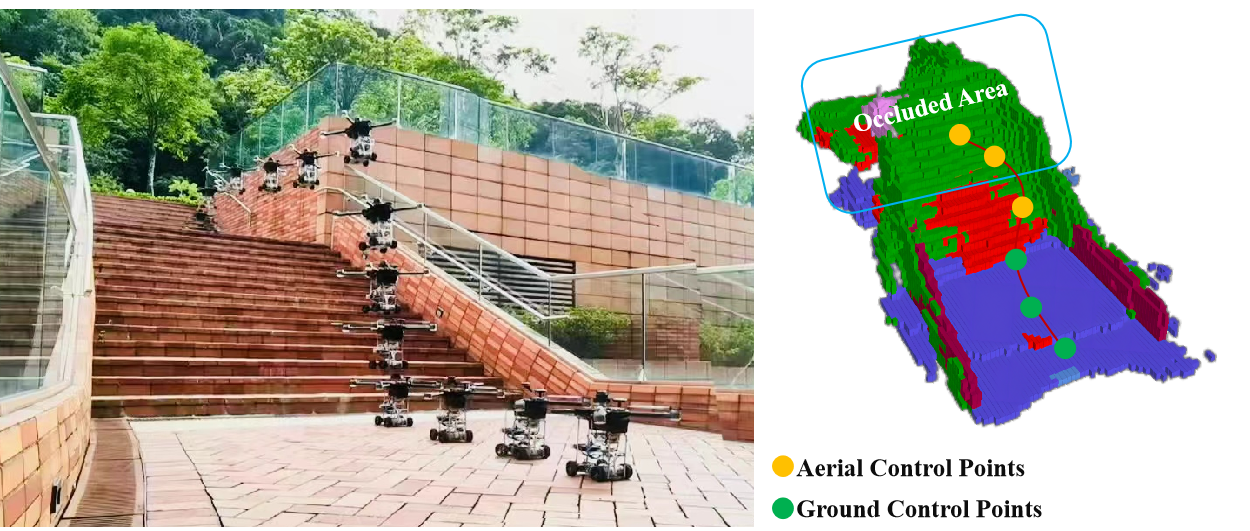}
    \caption{By predicting occlusions in advance, AGRNav~\cite{agrnav} gains an accurate perception of the environment and avoids collisions, resulting in efficient and energy-saving paths.}
    \label{fig:agrnav}
\end{figure}

\myparagraph{Workload}
We evaluated two typical real-world robotic applications on our testbed: Kapao, a real-time people-tracking application on our four-wheeled robot (Fig~\ref{fig:kapao}), and AGRNav, an autonomous navigation application on our air-ground robot (Fig~\ref{fig:agrnav}). 
These applications feature different model input and output size patterns: Kapao takes RGB images as input and outputs key points of small data volume. In contrast, AGRNav takes point clouds as input and outputs predicted point clouds and semantics of similar data volume as input, implying that AGRNav needs to transmit more data during distributed inference. 
And we have verified several models common to mobile devices on a larger scale to further corroborate our observations and findings: DenseNet~\cite{huang2018densely}, VGGNet~\cite{simonyan2015deep}, ConvNeXt~\cite{woo2023convnext}, RegNet~\cite{xu2022regnet}.

\myparagraph{Experiment Environments}
We evaluated two real-world environments: indoors (robots move in our laboratory with desks and separators interfering with wireless signals) and outdoors (robots move in our campus garden with trees and bushes interfering with wireless signals, resulting in lower bandwidth). 
The corresponding bandwidths between the robot and the GPU server in indoors and outdoors scenarios are shown in Fig.~\ref{fig:bandwidth}.

 

\myparagraph{Baselines}
We selected two SOTA pipeline parallelism methods as baselines: DSCCS ~\cite{liang2023dnn}, aimed at accelerating inference, and SPSO-GA ~\cite{chen2021energy}, focused on optimizing energy consumption. 
We set SPSO-GA's deadline constraints to 1 Hz, the minimum frequency required for robot movement control. 
Given our primary focus on inference time and energy consumption per inference, we disabled pipeline execution to concentrate solely on assessing the performance of various layer partitioning methods.

The evaluation questions are as follows:
\begin{itemize}
    \item RQ1: How does \xxx{} benefit real-world robotic applications compared to baseline systems in terms of inference time and energy consumption?
    \item RQ2: How sensitive is \xxx{} to bandwidth fluctuation in robotic IoT?
    \item RQ3: How does \xxx{} perform on models common to mobile devices on a larger scale?
    \item RQ4: What are the limitations and potentials of \xxx?
\end{itemize}

\subsection{Inference Time}



\begin{table*}[!t]
    \centering

\begin{tabular}{ccc|c|c|c|c|c|c}
\toprule
Model(number & Local compu- & \multirow[c]{2}{*}{System} & \multicolumn{2}{|c|}{Transmission time/s} & \multicolumn{2}{|c|}{Inference time/s} & \multicolumn{2}{c}{Percentage(\%)} \\
of parameters)& tation time/s &  & indoors & outdoors & indoors & outdoors & indoors & outdoors \\
\midrule
\multirow[c]{3}{*}{kapao(77M)} & \multirow[c]{3}{*}{1.01($\pm$0.03)} & SPSO-GA & 0.25($\pm$0.14) & 0.31($\pm$0.15) & 0.37($\pm$0.24) & 0.44($\pm$0.25) & 67.56 & 70.45 \\
 &  & DSCCS & 0.21($\pm$0.1) & 0.24($\pm$0.12) & 0.36($\pm$0.2) & 0.40($\pm$0.17) & 58.33 & 60.21 \\
 &  & Intra-DP & 0.24($\pm$0.15) & 0.28($\pm$0.13) & 0.31($\pm$0.14) & 0.34($\pm$0.12) & 77.42 & 82.35 \\
\cline{1-9} \cline{2-9}
\multirow[c]{3}{*}{agrnav(0.84M)} & \multirow[c]{3}{*}{0.60($\pm$0.04)} & SPSO-GA & 0.25($\pm$0.11) & 0.35($\pm$0.24) & 0.47($\pm$0.21) & 0.56($\pm$0.05) & 53.19 & 62.49 \\
 &  & DSCCS & 0.10($\pm$0.05) & 0.15($\pm$0.05) & 0.41($\pm$0.11) & 0.47($\pm$0.12) & 24.39 & 31.91\\
 &  & Intra-DP & 0.24($\pm$0.08) & 0.26($\pm$0.07) & 0.30($\pm$0.09) & 0.33($\pm$0.07) & 78.65 & 79.47 \\
\cline{1-9} \cline{2-9}
\bottomrule
\end{tabular}

    \caption{Average transmission time, inference time, percentage that transmission time accounts for of the total inference time and their standard deviation ($\pm n$) of Kapao and AGRNav in different environments with different systems. ``Local computation'' refers to inference the entire model locally on the robot.}
    \label{tab:e2e_time}
\end{table*}

\begin{table*}[!t]

\centering
\begin{tabular}{cc|c|c|c|c}
\toprule
 Model(number & \multirow[c]{2}{*}{System} & \multicolumn{2}{|c|}{Power consumption(W)} & \multicolumn{2}{|c}{Energy consumption(J) per inference} \\
 of parameters)&  & indoors & outdoors & indoors & outdoors \\
\midrule
\midrule
\multirow[c]{4}{*}{kapao(77M)} & Local & 10.61($\pm$0.49) & 10.61($\pm$0.49) & 9.79($\pm$0.03) & 9.79($\pm$0.03) \\
 & SPSO-GA & 5.49($\pm$1.52) & 5.35($\pm$1.37) & 2.03($\pm$0.82) & 2.35($\pm$1.04) \\
 & DSCCS & 6.38($\pm$2.21) & 6.63($\pm$2.38) & 2.30($\pm$0.55) & 2.65($\pm$0.55) \\
 & Intra-DP & 7.05($\pm$1.63) & 6.94($\pm$0.98) & 2.19($\pm$0.62) & 2.35($\pm$0.42) \\
\cline{1-6}
\multirow[c]{4}{*}{agrnav(0.84M)} & Local & 8.11($\pm$0.25) & 8.11($\pm$0.25) & 4.86($\pm$0.01) & 4.86($\pm$0.01) \\
 & SPSO-GA & 5.86($\pm$1.60) & 7.25($\pm$1.54) & 2.75($\pm$0.22) & 4.06($\pm$0.57) \\
 & DSCCS & 6.21($\pm$1.50) & 7.29($\pm$1.55) & 2.55($\pm$0.19) & 3.43($\pm$0.18) \\
 & Intra-DP & 7.52($\pm$0.51) & 8.04($\pm$0.45) & 2.26($\pm$0.15) & 2.63($\pm$0.15) \\
\cline{1-6}
\bottomrule
\end{tabular}

    \caption{The power consumption against time (Watt) and energy consumption per inference (Joule) with standard deviation ($\pm n$) of Kapao and AGRNav different environments with different systems. ``Local'' represents ``Local computation''.}
    \label{tab:e2e_power}
\end{table*}

The upper part of Tab.\ref{tab:e2e_time} demonstrates that \xxx{} significantly reduced Kapao's inference time compared to SPSO-GA and DSCCS, both indoors and outdoors. Specifically, \xxx{} achieved a 16.3\% and 14.9\% reduction indoors, and a 23.7\% and 15\% reduction outdoors, compared to SPSO-GA and DSCCS, respectively. 
For AGRNav, the performance gain of \xxx{} and baselines varied, as shown in the lower part of Tab.\ref{tab:e2e_time}. 
\xxx{} reduced AGRNav's inference time by 36.2\% and 27.8\% indoors, and 41.1\% and 30.8\% outdoors, compared to SPSO-GA and DSCCS, respectively.

Transmission time accounts for up to 70.45\% of the total inference time in SPSO-GA and DSCCS, highlighting the significant transmission bottlenecks faced by existing methods based on the PP paradigm, even with state-of-the-art layer partitioning. 
The difference between DSCCS and SPSO-GA can be attributed to their optimization goals: DSCCS minimizes inference latency, while SPSO-GA minimizes power consumption under deadline constraints. 
\xxx{}'s transmission time cannot reach close to 100\% of its inference time because it can only overlap the execution of local operators, and not all layers in the models of Kapao and AGRNav are local operator layers. 
Consequently, \xxx{} can only parallelize execution in some layers, not all layers.

The large standard deviation in transmission time outdoors for all systems indicates that bandwidth fluctuated more frequently and more fiercely outdoors compared to indoors, which is consistent with the observations in Fig.~\ref{fig:bandwidth}. Furthermore, the lower average bandwidth for outdoor scenarios (see Sec.~\ref{sec:bandwidth}) results in increased transmission and inference times relative to indoor scenarios.

\subsection{Energy Consumption}


\begin{table*}[htb]

    \centering
\begin{tabular}{ccc|c|c|c|c|c|c}
\toprule
 Model(number&  Local compu- & \multirow[c]{2}{*}{System} & \multicolumn{2}{|c|}{Transmission time/ms} & \multicolumn{2}{|c|}{Inference time/ms} & \multicolumn{2}{c}{Percentage(\%)} \\
 of parameters) & taion time/ms &  & indoors & outdoors & indoors & outdoors & indoors & outdoors \\
\midrule
\multirow[c]{3}{*}{DenseNet121(7M)} & \multirow[c]{3}{*}{74.5($\pm$18.7)} & SPSO-GA & 55.0($\pm$33.7) & 66.5($\pm$33.2) & 76.7($\pm$36.3) & 89.7($\pm$35.5) & 71.76 & 74.15 \\
 &  & DSCCS & 16.2($\pm$40.9) & 20.8($\pm$51.9) & 81.4($\pm$27.2) & 86.6($\pm$27.7) & 19.95 & 24.07 \\
 &  & Intra-DP & 53.4($\pm$34.5) & 52.9($\pm$23.9) & 74.5($\pm$850.7) & 55.1($\pm$15.6) & 71.70 & 96.05 \\
\cline{1-9} \cline{2-9}
\multirow[c]{3}{*}{RegNet(54M)} & \multirow[c]{3}{*}{175.0($\pm$23.6)} & SPSO-GA & 55.1($\pm$33.6) & 66.7($\pm$33.6) & 73.5($\pm$35.9) & 86.5($\pm$35.3) & 74.90 & 77.04 \\
 &  & DSCCS & 47.6($\pm$47.8) & 60.5($\pm$54.0) & 77.8($\pm$39.3) & 86.2($\pm$37.9) & 61.22 & 70.22 \\
 &  & Intra-DP & 49.6($\pm$21.7) & 59.9($\pm$23.4) & 55.0($\pm$24.8) & 64.2($\pm$25.2) & 90.18 & 93.34 \\
\cline{1-9} \cline{2-9}
\multirow[c]{3}{*}{ConvNeXt(88M)} & \multirow[c]{3}{*}{160.2($\pm$21.0)} & SPSO-GA & 55.4($\pm$33.9) & 66.9($\pm$34.9) & 73.8($\pm$35.4) & 86.7($\pm$36.3) & 75.13 & 77.15 \\
 &  & DSCCS & 46.9($\pm$43.1) & 56.7($\pm$52.1) & 72.4($\pm$35.7) & 84.7($\pm$36.3) & 64.78 & 66.95 \\
 &  & Intra-DP & 50.4($\pm$32.2) & 61.9($\pm$34.8) & 53.9($\pm$26.2) & 65.7($\pm$27.7) & 93.51 & 94.23 \\
\cline{1-9} \cline{2-9}
\multirow[c]{3}{*}{VGG19(143M)} & \multirow[c]{3}{*}{118.0($\pm$18.9)} & SPSO-GA & 55.7($\pm$33.5) & 67.2($\pm$35.0) & 68.8($\pm$33.9) & 80.6($\pm$35.0) & 80.84 & 83.33 \\
 &  & DSCCS & 38.9($\pm$47.1) & 41.6($\pm$53.8) & 65.2($\pm$28.1) & 75.5($\pm$27.1) & 59.75 & 55.09 \\
 &  & Intra-DP & 44.8($\pm$20.9) & 51.5($\pm$15.0) & 47.6($\pm$18.1) & 53.6($\pm$14.7) & 94.15 & 96.07 \\
\cline{1-9} \cline{2-9}
\multirow[c]{3}{*}{ConvNeXt(197M)} & \multirow[c]{3}{*}{316.7($\pm$31.0)} & SPSO-GA & 57.1($\pm$38.9) & 67.1($\pm$34.5) & 80.5($\pm$40.8) & 90.9($\pm$35.0) & 70.87 & 73.89 \\
 &  & DSCCS & 56.0($\pm$36.1) & 67.0($\pm$37.6) & 79.2($\pm$35.9) & 90.6($\pm$35.4) & 70.72 & 73.98 \\
 &  & Intra-DP & 56.4($\pm$34.7) & 66.5($\pm$33.7) & 59.7($\pm$26.6) & 68.0($\pm$26.6) & 94.43 & 97.88 \\
\cline{1-9} \cline{2-9}
\bottomrule
\end{tabular}

    \caption{Average transmission time, inference time, percentage that transmission time accounts for of the total inference time and their standard deviation ($\pm n$) of common AI models in different environments with different systems. }
    \label{tab:torchvision_time}
\end{table*}

\begin{table*}[htb]

\centering
\begin{tabular}{cc|c|c|c|c}
\toprule
Model(number & \multirow[c]{2}{*}{System} & \multicolumn{2}{|c|}{Power consumption(W)} & \multicolumn{2}{|c}{Energy consumption(J) per inference} \\
of parameters)&  & indoors & outdoors & indoors & outdoors \\
\midrule
\multirow[c]{4}{*}{DenseNet121(7M)} & Local & 8.2($\pm$0.27) & 8.2($\pm$0.27) & 0.46($\pm$0.04) & 0.46($\pm$0.04) \\
 & SPSO-GA & 4.87($\pm$0.23) & 4.74($\pm$0.19) & 0.37($\pm$0.02) & 0.42($\pm$0.02) \\
 & DSCCS & 6.91($\pm$0.45) & 6.86($\pm$0.46) & 0.56($\pm$0.04) & 0.59($\pm$0.04) \\
 & Intra-DP & 5.36($\pm$0.79) & 5.79($\pm$0.24) & 0.4($\pm$0.06) & 0.32($\pm$0.01) \\
\cline{1-6}
\multirow[c]{4}{*}{RegNet(54M)} & Local & 9.0($\pm$0.3) & 9.0($\pm$0.3) & 1.37($\pm$0.02) & 1.37($\pm$0.02) \\
 & SPSO-GA & 4.83($\pm$0.21) & 4.8($\pm$0.18) & 0.36($\pm$0.02) & 0.41($\pm$0.02) \\
 & DSCCS & 5.84($\pm$1.79) & 5.36($\pm$1.34) & 0.45($\pm$0.14) & 0.46($\pm$0.12) \\
 & Intra-DP & 5.24($\pm$1.43) & 5.28($\pm$1.52) & 0.29($\pm$0.08) & 0.34($\pm$0.1) \\
\cline{1-6}
\multirow[c]{4}{*}{ConvNeXt(88M)} & Local & 9.7($\pm$0.34) & 9.7($\pm$0.34) & 1.34($\pm$0.02) & 1.34($\pm$0.02) \\
 & SPSO-GA & 4.93($\pm$0.25) & 4.78($\pm$0.18) & 0.36($\pm$0.02) & 0.41($\pm$0.02) \\
 & DSCCS & 6.01($\pm$0.27) & 5.71($\pm$1.56) & 0.439($\pm$0.05) & 0.48($\pm$0.13) \\
 & Intra-DP & 6.68($\pm$1.23) & 6.68($\pm$1.21) & 0.36($\pm$0.07) & 0.44($\pm$0.08) \\
\cline{1-6}
\multirow[c]{4}{*}{VGG19(143M)} & Local & 9.78($\pm$0.34) & 9.78($\pm$0.34) & 0.95($\pm$0.02) & 0.95($\pm$0.02) \\
 & SPSO-GA & 4.9($\pm$0.25) & 4.82($\pm$0.2) & 0.34($\pm$0.02) & 0.39($\pm$0.02) \\
 & DSCCS & 6.58($\pm$2.14) & 6.93($\pm$2.35) & 0.43($\pm$0.14) & 0.52($\pm$0.18) \\
 & Intra-DP & 6.51($\pm$1.74) & 7.32($\pm$1.52) & 0.31($\pm$0.08) & 0.39($\pm$0.08) \\
\cline{1-6}
\multirow[c]{4}{*}{ConvNeXt(197M)} & Local & 10.72($\pm$0.38) & 10.72($\pm$0.38) & 3.12($\pm$0.03) & 3.12($\pm$0.03) \\
 & SPSO-GA & 5.1($\pm$0.27) & 4.99($\pm$0.2) & 0.41($\pm$0.02) & 0.45($\pm$0.02) \\
 & DSCCS & 5.06($\pm$0.31) & 5.02($\pm$0.37) & 0.4($\pm$0.02) & 0.45($\pm$0.03) \\
 & Intra-DP & 4.57($\pm$0.23) & 4.54($\pm$0.25) & 0.27($\pm$0.01) & 0.31($\pm$0.02) \\
\cline{1-6}
\bottomrule
\end{tabular}
    \caption{The power consumption against time (Watt) and energy consumption per inference (Joule) with standard deviation ($\pm n$) of common AI models in different environments with different systems. ``Local'' represents ``Local computation''.}
    \label{tab:torchvision_power}
\end{table*}

Table~\ref{tab:e2e_power} presents the power consumption over time and energy consumption per inference for Kapao and AGRNav using \xxx{} and baseline methods.
Compared to SPSO-GA and DSCCS, \xxx{} exhibits higher power consumption over time.
This can be attributed to \xxx{}'s computation at the operator granularity, where finer granularity results in lower GPU resource utilization and enables repeated computation of certain operators to avoids synchronization in \xxxschedule. 

Despite the higher power consumption over time, \xxx{} achieves the lowest energy consumption per inference for both Kapao and AGRNav, primarily due to its shortest inference time.
\xxx{} avoids the need for synchronization with high transmission costs in robotic IoT by introducing a small amount of redundant computation.
The additional energy consumed during \xxx{}'s computation phase is significantly lower than the energy wasted by the prolonged inference times of SPSO-GA and DSCCS.
Although SPSO-GA aims to optimize energy consumption, its advantages in power consumption over time diminish when considering energy consumption per inference due to extended inference times.
This is because SPSO-GA solely focuses on minimizing power consumption over time, potentially at the cost of prolonged inference time.





\subsection{Micro-Event Analysis}
\label{sec:breakdown}

\begin{figure}[htp]
    \centering
    \includegraphics[width=0.98\linewidth]{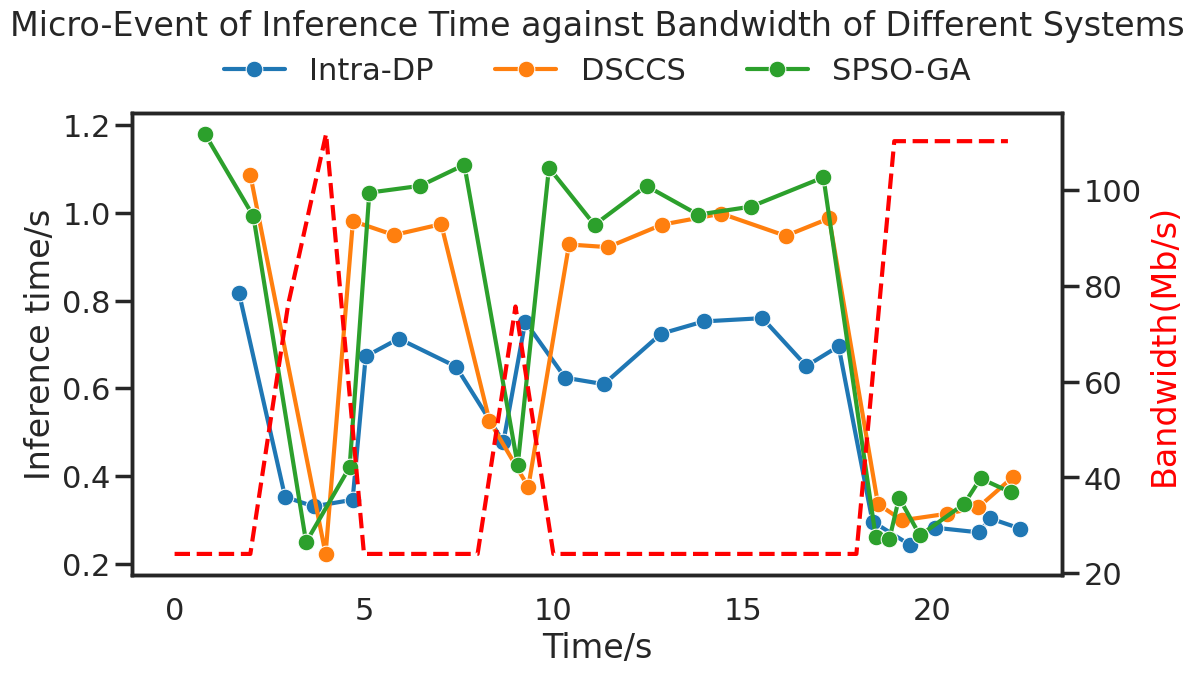}
    \caption{Real-time bandwidth and inference time of \xxx{} and baselines.}
    \label{fig:micro}
\end{figure}

To further investigate \xxx{}'s sensitivity to bandwidth fluctuations in robotic IoT, we recorded the real-time bandwidth and \xxx{}'s corresponding inference time response, as depicted in Fig.~\ref{fig:micro}.
When the bandwidth fluctuates, \xxx{}'s inference time also fluctuates due to the changing bandwidth.
However, the amplitude of inference time fluctuation is significantly smaller than that of bandwidth fluctuation, thanks to \xxx{}'s ability to flexibly switch between scheduling plans.
\xxx{} generates scheduling plans for different bandwidths based on possible bandwidth fluctuation ranges in advance and adopts the corresponding scheduling plan based on the predicted bandwidth during the inference phase.
This approach enables \xxx{} to adapt to varying network conditions and maintain stable performance, even in the presence of bandwidth fluctuations.

\subsection{Validation on a larger range of models}
We evaluated \xxx{} and baselines on a wide range of models commonly used in mobile devices, with parameter counts varying (detailed in Tab.~\ref{tab:torchvision_time} and Tab.~\ref{tab:torchvision_power}). 
Our results confirm that transmission time constitutes a significant portion of the total inference time in both DSCCS and SPSO-GA, leading to wasteful inference time and energy consumption compared to \xxx{}. 
Although \xxx{}'s outperformance remains consistent across various models, we observed that the performance gain is relatively smaller on models with fewer parameters. 
This is because \xxx{}'s performance improvement is primarily achieved through the parallel execution of local operators. 
When a model employs more global operator layers or has fewer parameters, the number of local operators available for parallel execution is reduced, limiting the optimization potential for \xxx{} to enhance performance.

\subsection{Lessons learned}
\myparagraph{Global optimal solution}
The performance of \xxx{} heavily depends on the quality of the solution obtained by \xxxschedule, with higher solution quality (closer to the global optimal solution) leading to better performance. 
However, as DNN models become increasingly complex with more layers, the nonlinear optimization problem in LOSS faces exponentially higher parameter dimensions and complexity, resulting in unacceptable profile times. 
Solving the global optimal solution of nonlinear optimization problems in finite time remains an open issue, and finding a fast and high-quality solution algorithm for \xxx{} is left for future work.

\myparagraph{Model structure}
During the implementation and evaluation of \xxx{}, we discovered that the presence of more local operator layers allows for increased parallel execution during model inference, thereby enhancing the performance improvement of \xxx{}. 
Future work should focus on supporting additional types of local operators and exploring the possibility of transforming global operators into local ones through lightweight synchronization techniques, based on their computational characteristics (e.g., synchronize the sum results in softmax instead of directly transferring the full input tensor and re-calculating).

\myparagraph{Future work}
It is of interest to explore further improvements of \xxx{}, such as a distributed inference system for multi-robot to minimize overall inference time and energy consumption. 
Such advancements could enable faster and more robust wireless distributed inference in real-world robotic IoT.

\section{Conclusion}
\label{sec:conclusion}
In this paper we present \xxx{}, a high-performance distributed inference system optimized for robotic IoT networks. 
By breaking up the granularity of model inference into local operators via \xxxparallel{} and applying adaptive scheduling to the computation and transmission of each local operator via \xxxschedule{}, \xxx{} dramatically reduces the transmission overhead in existing distributed inference on robotic IoT by overlapping the computation and transmission phases within the same inference task, achieving fast and energy-efficient distributed inference. 
We envision that the fast and energy-efficient inference of \xxx{} will foster the real-world deployment of diverse AI robotic tasks in the field.

\bibliographystyle{ACM-Reference-Format}
\bibliography{sample-base}


\begin{thebibliography}{65}


\ifx \showCODEN    \undefined \def \showCODEN     #1{\unskip}     \fi
\ifx \showDOI      \undefined \def \showDOI       #1{#1}\fi
\ifx \showISBNx    \undefined \def \showISBNx     #1{\unskip}     \fi
\ifx \showISBNxiii \undefined \def \showISBNxiii  #1{\unskip}     \fi
\ifx \showISSN     \undefined \def \showISSN      #1{\unskip}     \fi
\ifx \showLCCN     \undefined \def \showLCCN      #1{\unskip}     \fi
\ifx \shownote     \undefined \def \shownote      #1{#1}          \fi
\ifx \showarticletitle \undefined \def \showarticletitle #1{#1}   \fi
\ifx \showURL      \undefined \def \showURL       {\relax}        \fi
\providecommand\bibfield[2]{#2}
\providecommand\bibinfo[2]{#2}
\providecommand\natexlab[1]{#1}
\providecommand\showeprint[2][]{arXiv:#2}

\bibitem[noa({[n.\,d.]})]%
        {noauthor_iperf_nodate}
 \bibinfo{year}{[n.\,d.]}\natexlab{}.
\newblock \bibinfo{title}{{iPerf} - {Download} {iPerf3} and original {iPerf} pre-compiled binaries}.
\newblock
\newblock
\urldef\tempurl%
\url{https://iperf.fr/iperf-download.php}
\showURL{%
\tempurl}


\bibitem[Adame et~al\mbox{.}(2021)]%
        {adame2021time}
\bibfield{author}{\bibinfo{person}{Toni Adame}, \bibinfo{person}{Marc Carrascosa-Zamacois}, {and} \bibinfo{person}{Boris Bellalta}.} \bibinfo{year}{2021}\natexlab{}.
\newblock \showarticletitle{Time-sensitive networking in IEEE 802.11 be: On the way to low-latency WiFi 7}.
\newblock \bibinfo{journal}{\emph{Sensors}} \bibinfo{volume}{21}, \bibinfo{number}{15} (\bibinfo{year}{2021}), \bibinfo{pages}{4954}.
\newblock


\bibitem[Altamimi et~al\mbox{.}(2015)]%
        {altamimi2015energy}
\bibfield{author}{\bibinfo{person}{Majid Altamimi}, \bibinfo{person}{Atef Abdrabou}, \bibinfo{person}{Kshirasagar Naik}, {and} \bibinfo{person}{Amiya Nayak}.} \bibinfo{year}{2015}\natexlab{}.
\newblock \showarticletitle{Energy cost models of smartphones for task offloading to the cloud}.
\newblock \bibinfo{journal}{\emph{IEEE Transactions on Emerging Topics in Computing}} \bibinfo{volume}{3}, \bibinfo{number}{3} (\bibinfo{year}{2015}), \bibinfo{pages}{384--398}.
\newblock


\bibitem[Cao and de~Charette(2022)]%
        {cao2022monoscene}
\bibfield{author}{\bibinfo{person}{Anh-Quan Cao} {and} \bibinfo{person}{Raoul de Charette}.} \bibinfo{year}{2022}\natexlab{}.
\newblock \showarticletitle{Monoscene: Monocular 3d semantic scene completion}. In \bibinfo{booktitle}{\emph{Proceedings of the IEEE/CVF Conference on Computer Vision and Pattern Recognition}}. \bibinfo{pages}{3991--4001}.
\newblock


\bibitem[Chen et~al\mbox{.}(2021)]%
        {chen2021energy}
\bibfield{author}{\bibinfo{person}{Xing Chen}, \bibinfo{person}{Jianshan Zhang}, \bibinfo{person}{Bing Lin}, \bibinfo{person}{Zheyi Chen}, \bibinfo{person}{Katinka Wolter}, {and} \bibinfo{person}{Geyong Min}.} \bibinfo{year}{2021}\natexlab{}.
\newblock \showarticletitle{Energy-efficient offloading for DNN-based smart IoT systems in cloud-edge environments}.
\newblock \bibinfo{journal}{\emph{IEEE Transactions on Parallel and Distributed Systems}} \bibinfo{volume}{33}, \bibinfo{number}{3} (\bibinfo{year}{2021}), \bibinfo{pages}{683--697}.
\newblock


\bibitem[Crankshaw et~al\mbox{.}(2020)]%
        {crankshaw2020inferline}
\bibfield{author}{\bibinfo{person}{Daniel Crankshaw}, \bibinfo{person}{Gur-Eyal Sela}, \bibinfo{person}{Xiangxi Mo}, \bibinfo{person}{Corey Zumar}, \bibinfo{person}{Ion Stoica}, \bibinfo{person}{Joseph Gonzalez}, {and} \bibinfo{person}{Alexey Tumanov}.} \bibinfo{year}{2020}\natexlab{}.
\newblock \showarticletitle{InferLine: latency-aware provisioning and scaling for prediction serving pipelines}. In \bibinfo{booktitle}{\emph{Proceedings of the 11th ACM Symposium on Cloud Computing}}. \bibinfo{pages}{477--491}.
\newblock


\bibitem[Daubechies et~al\mbox{.}(2022)]%
        {daubechies2022nonlinear}
\bibfield{author}{\bibinfo{person}{Ingrid Daubechies}, \bibinfo{person}{Ronald DeVore}, \bibinfo{person}{Simon Foucart}, \bibinfo{person}{Boris Hanin}, {and} \bibinfo{person}{Guergana Petrova}.} \bibinfo{year}{2022}\natexlab{}.
\newblock \showarticletitle{Nonlinear approximation and (deep) ReLU networks}.
\newblock \bibinfo{journal}{\emph{Constructive Approximation}} \bibinfo{volume}{55}, \bibinfo{number}{1} (\bibinfo{year}{2022}), \bibinfo{pages}{127--172}.
\newblock


\bibitem[D{\'e}fossez et~al\mbox{.}(2021)]%
        {defossez2021differentiable}
\bibfield{author}{\bibinfo{person}{Alexandre D{\'e}fossez}, \bibinfo{person}{Yossi Adi}, {and} \bibinfo{person}{Gabriel Synnaeve}.} \bibinfo{year}{2021}\natexlab{}.
\newblock \showarticletitle{Differentiable model compression via pseudo quantization noise}.
\newblock \bibinfo{journal}{\emph{arXiv preprint arXiv:2104.09987}} (\bibinfo{year}{2021}).
\newblock


\bibitem[Ding et~al\mbox{.}(2015)]%
        {ding2015performance}
\bibfield{author}{\bibinfo{person}{Ming Ding}, \bibinfo{person}{Peng Wang}, \bibinfo{person}{David L{\'o}pez-P{\'e}rez}, \bibinfo{person}{Guoqiang Mao}, {and} \bibinfo{person}{Zihuai Lin}.} \bibinfo{year}{2015}\natexlab{}.
\newblock \showarticletitle{Performance impact of LoS and NLoS transmissions in dense cellular networks}.
\newblock \bibinfo{journal}{\emph{IEEE Transactions on Wireless Communications}} \bibinfo{volume}{15}, \bibinfo{number}{3} (\bibinfo{year}{2015}), \bibinfo{pages}{2365--2380}.
\newblock


\bibitem[Elgazzar et~al\mbox{.}(2014)]%
        {elgazzar2014cloud}
\bibfield{author}{\bibinfo{person}{Khalid Elgazzar}, \bibinfo{person}{Patrick Martin}, {and} \bibinfo{person}{Hossam~S Hassanein}.} \bibinfo{year}{2014}\natexlab{}.
\newblock \showarticletitle{Cloud-assisted computation offloading to support mobile services}.
\newblock \bibinfo{journal}{\emph{IEEE Transactions on Cloud Computing}} \bibinfo{volume}{4}, \bibinfo{number}{3} (\bibinfo{year}{2014}), \bibinfo{pages}{279--292}.
\newblock


\bibitem[Fang et~al\mbox{.}(2017)]%
        {fang2017qos}
\bibfield{author}{\bibinfo{person}{Zhou Fang}, \bibinfo{person}{Tong Yu}, \bibinfo{person}{Ole~J Mengshoel}, {and} \bibinfo{person}{Rajesh~K Gupta}.} \bibinfo{year}{2017}\natexlab{}.
\newblock \showarticletitle{Qos-aware scheduling of heterogeneous servers for inference in deep neural networks}. In \bibinfo{booktitle}{\emph{Proceedings of the 2017 ACM on Conference on Information and Knowledge Management}}. \bibinfo{pages}{2067--2070}.
\newblock


\bibitem[Fatahalian et~al\mbox{.}(2004)]%
        {fatahalian2004understanding}
\bibfield{author}{\bibinfo{person}{Kayvon Fatahalian}, \bibinfo{person}{Jeremy Sugerman}, {and} \bibinfo{person}{Pat Hanrahan}.} \bibinfo{year}{2004}\natexlab{}.
\newblock \showarticletitle{Understanding the efficiency of GPU algorithms for matrix-matrix multiplication}. In \bibinfo{booktitle}{\emph{Proceedings of the ACM SIGGRAPH/EUROGRAPHICS conference on Graphics hardware}}. \bibinfo{pages}{133--137}.
\newblock


\bibitem[Gheorghe and Ivanovici(2021)]%
        {gheorghe2021model}
\bibfield{author}{\bibinfo{person}{Ștefan Gheorghe} {and} \bibinfo{person}{Mihai Ivanovici}.} \bibinfo{year}{2021}\natexlab{}.
\newblock \showarticletitle{Model-based weight quantization for convolutional neural network compression}. In \bibinfo{booktitle}{\emph{2021 16th International Conference on Engineering of Modern Electric Systems (EMES)}}. IEEE, \bibinfo{pages}{1--4}.
\newblock


\bibitem[Gong et~al\mbox{.}(2020)]%
        {gong2020vecq}
\bibfield{author}{\bibinfo{person}{Cheng Gong}, \bibinfo{person}{Yao Chen}, \bibinfo{person}{Ye Lu}, \bibinfo{person}{Tao Li}, \bibinfo{person}{Cong Hao}, {and} \bibinfo{person}{Deming Chen}.} \bibinfo{year}{2020}\natexlab{}.
\newblock \showarticletitle{VecQ: Minimal loss DNN model compression with vectorized weight quantization}.
\newblock \bibinfo{journal}{\emph{IEEE Trans. Comput.}} \bibinfo{volume}{70}, \bibinfo{number}{5} (\bibinfo{year}{2020}), \bibinfo{pages}{696--710}.
\newblock


\bibitem[Gou et~al\mbox{.}(2021)]%
        {gou2021knowledge}
\bibfield{author}{\bibinfo{person}{Jianping Gou}, \bibinfo{person}{Baosheng Yu}, \bibinfo{person}{Stephen~J Maybank}, {and} \bibinfo{person}{Dacheng Tao}.} \bibinfo{year}{2021}\natexlab{}.
\newblock \showarticletitle{Knowledge distillation: A survey}.
\newblock \bibinfo{journal}{\emph{International Journal of Computer Vision}} \bibinfo{volume}{129}, \bibinfo{number}{6} (\bibinfo{year}{2021}), \bibinfo{pages}{1789--1819}.
\newblock


\bibitem[Honkote et~al\mbox{.}(2019)]%
        {honkote20192}
\bibfield{author}{\bibinfo{person}{Vinayak Honkote}, \bibinfo{person}{Dileep Kurian}, \bibinfo{person}{Sriram Muthukumar}, \bibinfo{person}{Dibyendu Ghosh}, \bibinfo{person}{Satish Yada}, \bibinfo{person}{Kartik Jain}, \bibinfo{person}{Bradley Jackson}, \bibinfo{person}{Ilya Klotchkov}, \bibinfo{person}{Mallikarjuna~Rao Nimmagadda}, \bibinfo{person}{Shreela Dattawadkar}, {et~al\mbox{.}}} \bibinfo{year}{2019}\natexlab{}.
\newblock \showarticletitle{2.4 a distributed autonomous and collaborative multi-robot system featuring a low-power robot soc in 22nm cmos for integrated battery-powered minibots}. In \bibinfo{booktitle}{\emph{2019 IEEE International Solid-State Circuits Conference-(ISSCC)}}. IEEE, \bibinfo{pages}{48--50}.
\newblock


\bibitem[Hu et~al\mbox{.}(2019)]%
        {hu2019dynamic}
\bibfield{author}{\bibinfo{person}{Chuang Hu}, \bibinfo{person}{Wei Bao}, \bibinfo{person}{Dan Wang}, {and} \bibinfo{person}{Fengming Liu}.} \bibinfo{year}{2019}\natexlab{}.
\newblock \showarticletitle{Dynamic adaptive DNN surgery for inference acceleration on the edge}. In \bibinfo{booktitle}{\emph{IEEE INFOCOM 2019-IEEE Conference on Computer Communications}}. IEEE, \bibinfo{pages}{1423--1431}.
\newblock


\bibitem[Hu et~al\mbox{.}(2022)]%
        {hu2022pipeedge}
\bibfield{author}{\bibinfo{person}{Yang Hu}, \bibinfo{person}{Connor Imes}, \bibinfo{person}{Xuanang Zhao}, \bibinfo{person}{Souvik Kundu}, \bibinfo{person}{Peter~A Beerel}, \bibinfo{person}{Stephen~P Crago}, {and} \bibinfo{person}{John~Paul Walters}.} \bibinfo{year}{2022}\natexlab{}.
\newblock \showarticletitle{Pipeedge: Pipeline parallelism for large-scale model inference on heterogeneous edge devices}. In \bibinfo{booktitle}{\emph{2022 25th Euromicro Conference on Digital System Design (DSD)}}. IEEE, \bibinfo{pages}{298--307}.
\newblock


\bibitem[Huang et~al\mbox{.}(2018)]%
        {huang2018densely}
\bibfield{author}{\bibinfo{person}{Gao Huang}, \bibinfo{person}{Zhuang Liu}, \bibinfo{person}{Laurens van~der Maaten}, {and} \bibinfo{person}{Kilian~Q. Weinberger}.} \bibinfo{year}{2018}\natexlab{}.
\newblock \bibinfo{title}{Densely Connected Convolutional Networks}.
\newblock
\newblock
\showeprint[arxiv]{1608.06993}~[cs.CV]


\bibitem[Jocher et~al\mbox{.}(2021)]%
        {jocher2021ultralytics}
\bibfield{author}{\bibinfo{person}{Glenn Jocher}, \bibinfo{person}{Alex Stoken}, \bibinfo{person}{Jirka Borovec}, \bibinfo{person}{Liu Changyu}, \bibinfo{person}{Adam Hogan}, \bibinfo{person}{Ayush Chaurasia}, \bibinfo{person}{Laurentiu Diaconu}, \bibinfo{person}{Francisco Ingham}, \bibinfo{person}{Adrien Colmagro}, \bibinfo{person}{Hu Ye}, {et~al\mbox{.}}} \bibinfo{year}{2021}\natexlab{}.
\newblock \showarticletitle{ultralytics/yolov5: v4. 0-nn. SiLU () activations, Weights \& Biases logging, PyTorch Hub integration}.
\newblock \bibinfo{journal}{\emph{Zenodo}} (\bibinfo{year}{2021}).
\newblock


\bibitem[Joseph et~al\mbox{.}(2021)]%
        {Joseph_2021_CVPR}
\bibfield{author}{\bibinfo{person}{K~J Joseph}, \bibinfo{person}{Salman Khan}, \bibinfo{person}{Fahad~Shahbaz Khan}, {and} \bibinfo{person}{Vineeth~N Balasubramanian}.} \bibinfo{year}{2021}\natexlab{}.
\newblock \showarticletitle{Towards Open World Object Detection}. In \bibinfo{booktitle}{\emph{Proceedings of the IEEE/CVF Conference on Computer Vision and Pattern Recognition (CVPR)}}. \bibinfo{pages}{5830--5840}.
\newblock


\bibitem[Kang et~al\mbox{.}(2017)]%
        {kang2017neurosurgeon}
\bibfield{author}{\bibinfo{person}{Yiping Kang}, \bibinfo{person}{Johann Hauswald}, \bibinfo{person}{Cao Gao}, \bibinfo{person}{Austin Rovinski}, \bibinfo{person}{Trevor Mudge}, \bibinfo{person}{Jason Mars}, {and} \bibinfo{person}{Lingjia Tang}.} \bibinfo{year}{2017}\natexlab{}.
\newblock \showarticletitle{Neurosurgeon: Collaborative intelligence between the cloud and mobile edge}.
\newblock \bibinfo{journal}{\emph{ACM SIGARCH Computer Architecture News}} \bibinfo{volume}{45}, \bibinfo{number}{1} (\bibinfo{year}{2017}), \bibinfo{pages}{615--629}.
\newblock


\bibitem[Kim et~al\mbox{.}(2003)]%
        {kim2003leakage}
\bibfield{author}{\bibinfo{person}{Nam~Sung Kim}, \bibinfo{person}{Todd Austin}, \bibinfo{person}{David Baauw}, \bibinfo{person}{Trevor Mudge}, \bibinfo{person}{Kriszti{\'a}n Flautner}, \bibinfo{person}{Jie~S Hu}, \bibinfo{person}{Mary~Jane Irwin}, \bibinfo{person}{Mahmut Kandemir}, {and} \bibinfo{person}{Vijaykrishnan Narayanan}.} \bibinfo{year}{2003}\natexlab{}.
\newblock \showarticletitle{Leakage current: Moore's law meets static power}.
\newblock \bibinfo{journal}{\emph{computer}} \bibinfo{volume}{36}, \bibinfo{number}{12} (\bibinfo{year}{2003}), \bibinfo{pages}{68--75}.
\newblock


\bibitem[Li et~al\mbox{.}(2019)]%
        {li2019evaluating}
\bibfield{author}{\bibinfo{person}{Ang Li}, \bibinfo{person}{Shuaiwen~Leon Song}, \bibinfo{person}{Jieyang Chen}, \bibinfo{person}{Jiajia Li}, \bibinfo{person}{Xu Liu}, \bibinfo{person}{Nathan~R Tallent}, {and} \bibinfo{person}{Kevin~J Barker}.} \bibinfo{year}{2019}\natexlab{}.
\newblock \showarticletitle{Evaluating modern gpu interconnect: Pcie, nvlink, nv-sli, nvswitch and gpudirect}.
\newblock \bibinfo{journal}{\emph{IEEE Transactions on Parallel and Distributed Systems}} \bibinfo{volume}{31}, \bibinfo{number}{1} (\bibinfo{year}{2019}), \bibinfo{pages}{94--110}.
\newblock


\bibitem[Li et~al\mbox{.}(2020)]%
        {li2020graph}
\bibfield{author}{\bibinfo{person}{Qingbiao Li}, \bibinfo{person}{Fernando Gama}, \bibinfo{person}{Alejandro Ribeiro}, {and} \bibinfo{person}{Amanda Prorok}.} \bibinfo{year}{2020}\natexlab{}.
\newblock \showarticletitle{Graph neural networks for decentralized multi-robot path planning}. In \bibinfo{booktitle}{\emph{2020 IEEE/RSJ International Conference on Intelligent Robots and Systems (IROS)}}. IEEE, \bibinfo{pages}{11785--11792}.
\newblock


\bibitem[Li et~al\mbox{.}(2023)]%
        {li2023voxformer}
\bibfield{author}{\bibinfo{person}{Yiming Li}, \bibinfo{person}{Zhiding Yu}, \bibinfo{person}{Christopher Choy}, \bibinfo{person}{Chaowei Xiao}, \bibinfo{person}{Jose~M Alvarez}, \bibinfo{person}{Sanja Fidler}, \bibinfo{person}{Chen Feng}, {and} \bibinfo{person}{Anima Anandkumar}.} \bibinfo{year}{2023}\natexlab{}.
\newblock \showarticletitle{Voxformer: Sparse voxel transformer for camera-based 3d semantic scene completion}. In \bibinfo{booktitle}{\emph{Proceedings of the IEEE/CVF Conference on Computer Vision and Pattern Recognition}}. \bibinfo{pages}{9087--9098}.
\newblock


\bibitem[Liang et~al\mbox{.}(2023)]%
        {liang2023dnn}
\bibfield{author}{\bibinfo{person}{Huanghuang Liang}, \bibinfo{person}{Qianlong Sang}, \bibinfo{person}{Chuang Hu}, \bibinfo{person}{Dazhao Cheng}, \bibinfo{person}{Xiaobo Zhou}, \bibinfo{person}{Dan Wang}, \bibinfo{person}{Wei Bao}, {and} \bibinfo{person}{Yu Wang}.} \bibinfo{year}{2023}\natexlab{}.
\newblock \showarticletitle{DNN surgery: Accelerating DNN inference on the edge through layer partitioning}.
\newblock \bibinfo{journal}{\emph{IEEE transactions on Cloud Computing}} (\bibinfo{year}{2023}).
\newblock


\bibitem[Lin et~al\mbox{.}(2019)]%
        {lin2019cost}
\bibfield{author}{\bibinfo{person}{Bing Lin}, \bibinfo{person}{Yinhao Huang}, \bibinfo{person}{Jianshan Zhang}, \bibinfo{person}{Junqin Hu}, \bibinfo{person}{Xing Chen}, {and} \bibinfo{person}{Jun Li}.} \bibinfo{year}{2019}\natexlab{}.
\newblock \showarticletitle{Cost-driven off-loading for DNN-based applications over cloud, edge, and end devices}.
\newblock \bibinfo{journal}{\emph{IEEE Transactions on Industrial Informatics}} \bibinfo{volume}{16}, \bibinfo{number}{8} (\bibinfo{year}{2019}), \bibinfo{pages}{5456--5466}.
\newblock


\bibitem[Lin et~al\mbox{.}(2020)]%
        {lin2020ensemble}
\bibfield{author}{\bibinfo{person}{Tao Lin}, \bibinfo{person}{Lingjing Kong}, \bibinfo{person}{Sebastian~U Stich}, {and} \bibinfo{person}{Martin Jaggi}.} \bibinfo{year}{2020}\natexlab{}.
\newblock \showarticletitle{Ensemble distillation for robust model fusion in federated learning}.
\newblock \bibinfo{journal}{\emph{Advances in Neural Information Processing Systems}}  \bibinfo{volume}{33} (\bibinfo{year}{2020}), \bibinfo{pages}{2351--2363}.
\newblock


\bibitem[Liu and Choi(2023)]%
        {liu2023first}
\bibfield{author}{\bibinfo{person}{Ruofeng Liu} {and} \bibinfo{person}{Nakjung Choi}.} \bibinfo{year}{2023}\natexlab{}.
\newblock \showarticletitle{A First Look at Wi-Fi 6 in Action: Throughput, Latency, Energy Efficiency, and Security}.
\newblock \bibinfo{journal}{\emph{Proceedings of the ACM on Measurement and Analysis of Computing Systems}} \bibinfo{volume}{7}, \bibinfo{number}{1} (\bibinfo{year}{2023}), \bibinfo{pages}{1--25}.
\newblock


\bibitem[Liu et~al\mbox{.}(2022)]%
        {Liu_2022_CVPR}
\bibfield{author}{\bibinfo{person}{Shuai Liu}, \bibinfo{person}{Xin Li}, \bibinfo{person}{Huchuan Lu}, {and} \bibinfo{person}{You He}.} \bibinfo{year}{2022}\natexlab{}.
\newblock \showarticletitle{Multi-Object Tracking Meets Moving UAV}. In \bibinfo{booktitle}{\emph{Proceedings of the IEEE/CVF Conference on Computer Vision and Pattern Recognition (CVPR)}}. \bibinfo{pages}{8876--8885}.
\newblock


\bibitem[Liu et~al\mbox{.}(2016)]%
        {liu2016large}
\bibfield{author}{\bibinfo{person}{Weiyang Liu}, \bibinfo{person}{Yandong Wen}, \bibinfo{person}{Zhiding Yu}, {and} \bibinfo{person}{Meng Yang}.} \bibinfo{year}{2016}\natexlab{}.
\newblock \showarticletitle{Large-margin softmax loss for convolutional neural networks}.
\newblock \bibinfo{journal}{\emph{arXiv preprint arXiv:1612.02295}} (\bibinfo{year}{2016}).
\newblock


\bibitem[Masiukiewicz(2019)]%
        {masiukiewicz2019throughput}
\bibfield{author}{\bibinfo{person}{Antoni Masiukiewicz}.} \bibinfo{year}{2019}\natexlab{}.
\newblock \showarticletitle{Throughput comparison between the new HEW 802.11 ax standard and 802.11 n/ac standards in selected distance windows}.
\newblock \bibinfo{journal}{\emph{International Journal of Electronics and Telecommunications}} \bibinfo{volume}{65}, \bibinfo{number}{1} (\bibinfo{year}{2019}), \bibinfo{pages}{79--84}.
\newblock


\bibitem[McNally et~al\mbox{.}(2022)]%
        {kapao}
\bibfield{author}{\bibinfo{person}{William McNally}, \bibinfo{person}{Kanav Vats}, \bibinfo{person}{Alexander Wong}, {and} \bibinfo{person}{John McPhee}.} \bibinfo{year}{2022}\natexlab{}.
\newblock \showarticletitle{Rethinking keypoint representations: Modeling keypoints and poses as objects for multi-person human pose estimation}. In \bibinfo{booktitle}{\emph{European Conference on Computer Vision}}. Springer, \bibinfo{pages}{37--54}.
\newblock


\bibitem[Mohammed et~al\mbox{.}(2020)]%
        {mohammed2020distributed}
\bibfield{author}{\bibinfo{person}{Thaha Mohammed}, \bibinfo{person}{Carlee Joe-Wong}, \bibinfo{person}{Rohit Babbar}, {and} \bibinfo{person}{Mario Di~Francesco}.} \bibinfo{year}{2020}\natexlab{}.
\newblock \showarticletitle{Distributed inference acceleration with adaptive DNN partitioning and offloading}. In \bibinfo{booktitle}{\emph{IEEE INFOCOM 2020-IEEE Conference on Computer Communications}}. IEEE, \bibinfo{pages}{854--863}.
\newblock


\bibitem[Narayanan et~al\mbox{.}(2021)]%
        {narayanan2021efficient}
\bibfield{author}{\bibinfo{person}{Deepak Narayanan}, \bibinfo{person}{Mohammad Shoeybi}, \bibinfo{person}{Jared Casper}, \bibinfo{person}{Patrick LeGresley}, \bibinfo{person}{Mostofa Patwary}, \bibinfo{person}{Vijay Korthikanti}, \bibinfo{person}{Dmitri Vainbrand}, \bibinfo{person}{Prethvi Kashinkunti}, \bibinfo{person}{Julie Bernauer}, \bibinfo{person}{Bryan Catanzaro}, {et~al\mbox{.}}} \bibinfo{year}{2021}\natexlab{}.
\newblock \showarticletitle{Efficient large-scale language model training on gpu clusters using megatron-lm}. In \bibinfo{booktitle}{\emph{Proceedings of the International Conference for High Performance Computing, Networking, Storage and Analysis}}. \bibinfo{pages}{1--15}.
\newblock


\bibitem[Noormohammadpour and Raghavendra(2017)]%
        {noormohammadpour2017datacenter}
\bibfield{author}{\bibinfo{person}{Mohammad Noormohammadpour} {and} \bibinfo{person}{Cauligi~S Raghavendra}.} \bibinfo{year}{2017}\natexlab{}.
\newblock \showarticletitle{Datacenter traffic control: Understanding techniques and tradeoffs}.
\newblock \bibinfo{journal}{\emph{IEEE Communications Surveys \& Tutorials}} \bibinfo{volume}{20}, \bibinfo{number}{2} (\bibinfo{year}{2017}), \bibinfo{pages}{1492--1525}.
\newblock


\bibitem[NVIDIA(2024)]%
        {jetsonnx}
\bibfield{author}{\bibinfo{person}{NVIDIA}.} \bibinfo{year}{2024}\natexlab{}.
\newblock \bibinfo{title}{The World's Smallest AI Supercomputer}.
\newblock \bibinfo{howpublished}{\url{https://www.nvidia.com/en-us/autonomous-machines/embedded-systems/jetson-xavier-series/}}.
\newblock


\bibitem[Ohkawa et~al\mbox{.}(2018)]%
        {ohkawa2018fpga}
\bibfield{author}{\bibinfo{person}{Takeshi Ohkawa}, \bibinfo{person}{Kazushi Yamashina}, \bibinfo{person}{Hitomi Kimura}, \bibinfo{person}{Kanemitsu Ootsu}, {and} \bibinfo{person}{Takashi Yokota}.} \bibinfo{year}{2018}\natexlab{}.
\newblock \showarticletitle{FPGA components for integrating FPGAs into robot systems}.
\newblock \bibinfo{journal}{\emph{IEICE TRANSACTIONS on Information and Systems}} \bibinfo{volume}{101}, \bibinfo{number}{2} (\bibinfo{year}{2018}), \bibinfo{pages}{363--375}.
\newblock


\bibitem[Pei et~al\mbox{.}(2013)]%
        {pei2013connectivity}
\bibfield{author}{\bibinfo{person}{Yuanteng Pei}, \bibinfo{person}{Matt~W Mutka}, {and} \bibinfo{person}{Ning Xi}.} \bibinfo{year}{2013}\natexlab{}.
\newblock \showarticletitle{Connectivity and bandwidth-aware real-time exploration in mobile robot networks}.
\newblock \bibinfo{journal}{\emph{Wireless Communications and Mobile Computing}} \bibinfo{volume}{13}, \bibinfo{number}{9} (\bibinfo{year}{2013}), \bibinfo{pages}{847--863}.
\newblock


\bibitem[pytorch(2024a)]%
        {pytorch}
\bibfield{author}{\bibinfo{person}{pytorch}.} \bibinfo{year}{2024}\natexlab{a}.
\newblock \bibinfo{title}{pytroch}.
\newblock \bibinfo{howpublished}{\url{https://pytorch.org/}}.
\newblock


\bibitem[pytorch(2024b)]%
        {conv2d}
\bibfield{author}{\bibinfo{person}{pytorch}.} \bibinfo{year}{2024}\natexlab{b}.
\newblock \bibinfo{title}{pytroch}.
\newblock \bibinfo{howpublished}{\url{https://pytorch.org/docs/stable/generated/torch.nn.Conv2d.html}}.
\newblock


\bibitem[Qin et~al\mbox{.}(2008)]%
        {qin2008differential}
\bibfield{author}{\bibinfo{person}{A~Kai Qin}, \bibinfo{person}{Vicky~Ling Huang}, {and} \bibinfo{person}{Ponnuthurai~N Suganthan}.} \bibinfo{year}{2008}\natexlab{}.
\newblock \showarticletitle{Differential evolution algorithm with strategy adaptation for global numerical optimization}.
\newblock \bibinfo{journal}{\emph{IEEE transactions on Evolutionary Computation}} \bibinfo{volume}{13}, \bibinfo{number}{2} (\bibinfo{year}{2008}), \bibinfo{pages}{398--417}.
\newblock


\bibitem[Ren et~al\mbox{.}(2018)]%
        {ren2018proportional}
\bibfield{author}{\bibinfo{person}{Yi Ren}, \bibinfo{person}{Chih-Wei Tung}, \bibinfo{person}{Jyh-Cheng Chen}, {and} \bibinfo{person}{Frank~Y Li}.} \bibinfo{year}{2018}\natexlab{}.
\newblock \showarticletitle{Proportional and preemption-enabled traffic offloading for IP flow mobility: Algorithms and performance evaluation}.
\newblock \bibinfo{journal}{\emph{IEEE Transactions on Vehicular Technology}} \bibinfo{volume}{67}, \bibinfo{number}{12} (\bibinfo{year}{2018}), \bibinfo{pages}{12095--12108}.
\newblock


\bibitem[Sarkar and Mussa(2013)]%
        {sarkar2013effect}
\bibfield{author}{\bibinfo{person}{Nurul~I Sarkar} {and} \bibinfo{person}{Osman Mussa}.} \bibinfo{year}{2013}\natexlab{}.
\newblock \showarticletitle{The effect of people movement on Wi-Fi link throughput in indoor propagation environments}. In \bibinfo{booktitle}{\emph{IEEE 2013 Tencon-Spring}}. IEEE, \bibinfo{pages}{562--566}.
\newblock


\bibitem[Simonyan and Zisserman(2015)]%
        {simonyan2015deep}
\bibfield{author}{\bibinfo{person}{Karen Simonyan} {and} \bibinfo{person}{Andrew Zisserman}.} \bibinfo{year}{2015}\natexlab{}.
\newblock \bibinfo{title}{Very Deep Convolutional Networks for Large-Scale Image Recognition}.
\newblock
\newblock
\showeprint[arxiv]{1409.1556}~[cs.CV]


\bibitem[Sinha and El-Sharkawy(2019)]%
        {sinha2019thin}
\bibfield{author}{\bibinfo{person}{Debjyoti Sinha} {and} \bibinfo{person}{Mohamed El-Sharkawy}.} \bibinfo{year}{2019}\natexlab{}.
\newblock \showarticletitle{Thin mobilenet: An enhanced mobilenet architecture}. In \bibinfo{booktitle}{\emph{2019 IEEE 10th annual ubiquitous computing, electronics \& mobile communication conference (UEMCON)}}. IEEE, \bibinfo{pages}{0280--0285}.
\newblock


\bibitem[Sun et~al\mbox{.}(2021)]%
        {sun2021ampnet}
\bibfield{author}{\bibinfo{person}{Luna Sun}, \bibinfo{person}{Zhenxue Chen}, \bibinfo{person}{QM~Jonathan Wu}, \bibinfo{person}{Hongjian Zhao}, \bibinfo{person}{Weikai He}, {and} \bibinfo{person}{Xinghe Yan}.} \bibinfo{year}{2021}\natexlab{}.
\newblock \showarticletitle{AMPNet: Average-and max-pool networks for salient object detection}.
\newblock \bibinfo{journal}{\emph{IEEE Transactions on Circuits and Systems for Video Technology}} \bibinfo{volume}{31}, \bibinfo{number}{11} (\bibinfo{year}{2021}), \bibinfo{pages}{4321--4333}.
\newblock


\bibitem[Targ et~al\mbox{.}(2016)]%
        {targ2016resnet}
\bibfield{author}{\bibinfo{person}{Sasha Targ}, \bibinfo{person}{Diogo Almeida}, {and} \bibinfo{person}{Kevin Lyman}.} \bibinfo{year}{2016}\natexlab{}.
\newblock \showarticletitle{Resnet in resnet: Generalizing residual architectures}.
\newblock \bibinfo{journal}{\emph{arXiv preprint arXiv:1603.08029}} (\bibinfo{year}{2016}).
\newblock


\bibitem[Wang et~al\mbox{.}(2011)]%
        {wang2011mvapich2}
\bibfield{author}{\bibinfo{person}{Hao Wang}, \bibinfo{person}{Sreeram Potluri}, \bibinfo{person}{Miao Luo}, \bibinfo{person}{Ashish~Kumar Singh}, \bibinfo{person}{Sayantan Sur}, {and} \bibinfo{person}{Dhabaleswar~K Panda}.} \bibinfo{year}{2011}\natexlab{}.
\newblock \showarticletitle{MVAPICH2-GPU: optimized GPU to GPU communication for InfiniBand clusters}.
\newblock \bibinfo{journal}{\emph{Computer Science-Research and Development}} \bibinfo{volume}{26}, \bibinfo{number}{3} (\bibinfo{year}{2011}), \bibinfo{pages}{257--266}.
\newblock


\bibitem[Wang et~al\mbox{.}(2024)]%
        {agrnav}
\bibfield{author}{\bibinfo{person}{Junming Wang}, \bibinfo{person}{Zekai Sun}, \bibinfo{person}{Xiuxian Guan}, \bibinfo{person}{Tianxiang Shen}, \bibinfo{person}{Zongyuan Zhang}, \bibinfo{person}{Tianyang Duan}, \bibinfo{person}{Dong Huang}, \bibinfo{person}{Shixiong Zhao}, {and} \bibinfo{person}{Heming Cui}.} \bibinfo{year}{2024}\natexlab{}.
\newblock \showarticletitle{AGRNav: Efficient and Energy-Saving Autonomous Navigation for Air-Ground Robots in Occlusion-Prone Environments}. In \bibinfo{booktitle}{\emph{IEEE International Conference on Robotics and Automation (ICRA)}}.
\newblock


\bibitem[Wang and Yoon(2021)]%
        {wang2021knowledge}
\bibfield{author}{\bibinfo{person}{Lin Wang} {and} \bibinfo{person}{Kuk-Jin Yoon}.} \bibinfo{year}{2021}\natexlab{}.
\newblock \showarticletitle{Knowledge distillation and student-teacher learning for visual intelligence: A review and new outlooks}.
\newblock \bibinfo{journal}{\emph{IEEE transactions on pattern analysis and machine intelligence}} \bibinfo{volume}{44}, \bibinfo{number}{6} (\bibinfo{year}{2021}), \bibinfo{pages}{3048--3068}.
\newblock


\bibitem[Woo et~al\mbox{.}(2023)]%
        {woo2023convnext}
\bibfield{author}{\bibinfo{person}{Sanghyun Woo}, \bibinfo{person}{Shoubhik Debnath}, \bibinfo{person}{Ronghang Hu}, \bibinfo{person}{Xinlei Chen}, \bibinfo{person}{Zhuang Liu}, \bibinfo{person}{In~So Kweon}, {and} \bibinfo{person}{Saining Xie}.} \bibinfo{year}{2023}\natexlab{}.
\newblock \showarticletitle{Convnext v2: Co-designing and scaling convnets with masked autoencoders}. In \bibinfo{booktitle}{\emph{Proceedings of the IEEE/CVF Conference on Computer Vision and Pattern Recognition}}. \bibinfo{pages}{16133--16142}.
\newblock


\bibitem[Wu et~al\mbox{.}(2019)]%
        {wu2019efficient}
\bibfield{author}{\bibinfo{person}{Huaming Wu}, \bibinfo{person}{William~J Knottenbelt}, {and} \bibinfo{person}{Katinka Wolter}.} \bibinfo{year}{2019}\natexlab{}.
\newblock \showarticletitle{An efficient application partitioning algorithm in mobile environments}.
\newblock \bibinfo{journal}{\emph{IEEE Transactions on Parallel and Distributed Systems}} \bibinfo{volume}{30}, \bibinfo{number}{7} (\bibinfo{year}{2019}), \bibinfo{pages}{1464--1480}.
\newblock


\bibitem[Wu et~al\mbox{.}(2023)]%
        {wu2023brief}
\bibfield{author}{\bibinfo{person}{Tianyu Wu}, \bibinfo{person}{Shizhu He}, \bibinfo{person}{Jingping Liu}, \bibinfo{person}{Siqi Sun}, \bibinfo{person}{Kang Liu}, \bibinfo{person}{Qing-Long Han}, {and} \bibinfo{person}{Yang Tang}.} \bibinfo{year}{2023}\natexlab{}.
\newblock \showarticletitle{A brief overview of ChatGPT: The history, status quo and potential future development}.
\newblock \bibinfo{journal}{\emph{IEEE/CAA Journal of Automatica Sinica}} \bibinfo{volume}{10}, \bibinfo{number}{5} (\bibinfo{year}{2023}), \bibinfo{pages}{1122--1136}.
\newblock


\bibitem[Xia et~al\mbox{.}(2023)]%
        {xia2023scpnet}
\bibfield{author}{\bibinfo{person}{Zhaoyang Xia}, \bibinfo{person}{Youquan Liu}, \bibinfo{person}{Xin Li}, \bibinfo{person}{Xinge Zhu}, \bibinfo{person}{Yuexin Ma}, \bibinfo{person}{Yikang Li}, \bibinfo{person}{Yuenan Hou}, {and} \bibinfo{person}{Yu Qiao}.} \bibinfo{year}{2023}\natexlab{}.
\newblock \showarticletitle{SCPNet: Semantic Scene Completion on Point Cloud}. In \bibinfo{booktitle}{\emph{Proceedings of the IEEE/CVF Conference on Computer Vision and Pattern Recognition}}. \bibinfo{pages}{17642--17651}.
\newblock


\bibitem[Xiang and Kim(2019)]%
        {xiang2019pipelined}
\bibfield{author}{\bibinfo{person}{Yecheng Xiang} {and} \bibinfo{person}{Hyoseung Kim}.} \bibinfo{year}{2019}\natexlab{}.
\newblock \showarticletitle{Pipelined data-parallel CPU/GPU scheduling for multi-DNN real-time inference}. In \bibinfo{booktitle}{\emph{2019 IEEE Real-Time Systems Symposium (RTSS)}}. IEEE, \bibinfo{pages}{392--405}.
\newblock


\bibitem[Xu et~al\mbox{.}(2022)]%
        {xu2022regnet}
\bibfield{author}{\bibinfo{person}{Jing Xu}, \bibinfo{person}{Yu Pan}, \bibinfo{person}{Xinglin Pan}, \bibinfo{person}{Steven Hoi}, \bibinfo{person}{Zhang Yi}, {and} \bibinfo{person}{Zenglin Xu}.} \bibinfo{year}{2022}\natexlab{}.
\newblock \showarticletitle{RegNet: self-regulated network for image classification}.
\newblock \bibinfo{journal}{\emph{IEEE Transactions on Neural Networks and Learning Systems}} (\bibinfo{year}{2022}).
\newblock


\bibitem[Xue et~al\mbox{.}(2021)]%
        {xue2021ddpqn}
\bibfield{author}{\bibinfo{person}{Min Xue}, \bibinfo{person}{Huaming Wu}, \bibinfo{person}{Guang Peng}, {and} \bibinfo{person}{Katinka Wolter}.} \bibinfo{year}{2021}\natexlab{}.
\newblock \showarticletitle{DDPQN: An efficient DNN offloading strategy in local-edge-cloud collaborative environments}.
\newblock \bibinfo{journal}{\emph{IEEE Transactions on Services Computing}} \bibinfo{volume}{15}, \bibinfo{number}{2} (\bibinfo{year}{2021}), \bibinfo{pages}{640--655}.
\newblock


\bibitem[Yang et~al\mbox{.}(2022)]%
        {yang2022mobile}
\bibfield{author}{\bibinfo{person}{Xinlei Yang}, \bibinfo{person}{Hao Lin}, \bibinfo{person}{Zhenhua Li}, \bibinfo{person}{Feng Qian}, \bibinfo{person}{Xingyao Li}, \bibinfo{person}{Zhiming He}, \bibinfo{person}{Xudong Wu}, \bibinfo{person}{Xianlong Wang}, \bibinfo{person}{Yunhao Liu}, \bibinfo{person}{Zhi Liao}, {et~al\mbox{.}}} \bibinfo{year}{2022}\natexlab{}.
\newblock \showarticletitle{Mobile access bandwidth in practice: Measurement, analysis, and implications}. In \bibinfo{booktitle}{\emph{Proceedings of the ACM SIGCOMM 2022 Conference}}. \bibinfo{pages}{114--128}.
\newblock


\bibitem[Yang et~al\mbox{.}(2020)]%
        {yang2020multi}
\bibfield{author}{\bibinfo{person}{Yang Yang}, \bibinfo{person}{Li Juntao}, {and} \bibinfo{person}{Peng Lingling}.} \bibinfo{year}{2020}\natexlab{}.
\newblock \showarticletitle{Multi-robot path planning based on a deep reinforcement learning DQN algorithm}.
\newblock \bibinfo{journal}{\emph{CAAI Transactions on Intelligence Technology}} \bibinfo{volume}{5}, \bibinfo{number}{3} (\bibinfo{year}{2020}), \bibinfo{pages}{177--183}.
\newblock


\bibitem[Yin et~al\mbox{.}(2003)]%
        {yin2003flexible}
\bibfield{author}{\bibinfo{person}{Xinyou Yin}, \bibinfo{person}{JAN Goudriaan}, \bibinfo{person}{Egbert~A Lantinga}, \bibinfo{person}{JAN Vos}, {and} \bibinfo{person}{Huub~J Spiertz}.} \bibinfo{year}{2003}\natexlab{}.
\newblock \showarticletitle{A flexible sigmoid function of determinate growth}.
\newblock \bibinfo{journal}{\emph{Annals of botany}} \bibinfo{volume}{91}, \bibinfo{number}{3} (\bibinfo{year}{2003}), \bibinfo{pages}{361--371}.
\newblock


\bibitem[Yue et~al\mbox{.}(2017)]%
        {yue2017linkforecast}
\bibfield{author}{\bibinfo{person}{Chaoqun Yue}, \bibinfo{person}{Ruofan Jin}, \bibinfo{person}{Kyoungwon Suh}, \bibinfo{person}{Yanyuan Qin}, \bibinfo{person}{Bing Wang}, {and} \bibinfo{person}{Wei Wei}.} \bibinfo{year}{2017}\natexlab{}.
\newblock \showarticletitle{LinkForecast: Cellular link bandwidth prediction in LTE networks}.
\newblock \bibinfo{journal}{\emph{IEEE Transactions on Mobile Computing}} \bibinfo{volume}{17}, \bibinfo{number}{7} (\bibinfo{year}{2017}), \bibinfo{pages}{1582--1594}.
\newblock


\bibitem[Zee(1996)]%
        {zee1996law}
\bibfield{author}{\bibinfo{person}{Anthony Zee}.} \bibinfo{year}{1996}\natexlab{}.
\newblock \showarticletitle{Law of addition in random matrix theory}.
\newblock \bibinfo{journal}{\emph{Nuclear Physics B}} \bibinfo{volume}{474}, \bibinfo{number}{3} (\bibinfo{year}{1996}), \bibinfo{pages}{726--744}.
\newblock


\bibitem[Zhuang et~al\mbox{.}(2023)]%
        {zhuang2023optimizing}
\bibfield{author}{\bibinfo{person}{Yonghao Zhuang}, \bibinfo{person}{Hexu Zhao}, \bibinfo{person}{Lianmin Zheng}, \bibinfo{person}{Zhuohan Li}, \bibinfo{person}{Eric Xing}, \bibinfo{person}{Qirong Ho}, \bibinfo{person}{Joseph Gonzalez}, \bibinfo{person}{Ion Stoica}, {and} \bibinfo{person}{Hao Zhang}.} \bibinfo{year}{2023}\natexlab{}.
\newblock \showarticletitle{On optimizing the communication of model parallelism}.
\newblock \bibinfo{journal}{\emph{Proceedings of Machine Learning and Systems}}  \bibinfo{volume}{5} (\bibinfo{year}{2023}).
\newblock


\end{thebibliography}










\end{document}